\documentclass[journal]{IEEEtranTIE}
\usepackage{graphicx}
\graphicspath{{./figures/}}
\usepackage{cite}
\usepackage{picinpar}
\usepackage{amsmath, amsfonts}
\usepackage[caption=false,font=footnotesize]{subfig}
\usepackage{url}
\usepackage{flushend}
\usepackage[latin1]{inputenc}
\usepackage{colortbl}
\usepackage{soul}
\usepackage{multirow}
\usepackage{pifont}
\usepackage{color}
\usepackage{alltt}
\usepackage[hidelinks]{hyperref}
\usepackage{enumerate}
\usepackage{siunitx}
\usepackage{breakurl}
\usepackage{epstopdf}
\usepackage{pbox}

\begin{document}
\title{Fault Diagnosis in Microelectronics Attachment via Deep Learning Analysis of 3D Laser Scans}

\author{
%	\vskip 1em
	Nikolaos~Dimitriou,
  Lampros~Leontaris,
  Thanasis~Vafeiadis,
  Dimosthenis~Ioannidis,
  Tracy~Wotherspoon,
  Gregory~Tinker,
  and~Dimitrios~Tzovaras
\thanks{This work has been partially supported by the European Commission through project Z-Fact0r funded by the European Union H2020 programme under Grant Agreement no. 723906. The opinions expressed in this paper are those of the authors and do not necessarily reflect the views of the European Commission.}
\thanks{Nikolaos Dimitriou, Lampros~Leontaris, Thanasis Vafeiadis, Dimosthenis Ioannidis and Dimitrios Tzovaras are  with  the  Information  Technologies  Institute,  Centre  for Research  and Technology  Hellas  (CERTH-ITI), 57001  Thessaloniki,  Greece. (e-mail: nikdim@iti.gr, lleontar@iti.gr, thanvaf@iti.gr, djoannid@iti.gr, dimitrios.tzovaras@iti.gr).}
\thanks{Tracy Wotherspoon and Gregory Tinker are  with  Microsemi Corporation, a wholly owned subsidiary of Microchip Technology. (e-mail: Tracy.Wotherspoon@microchip.com, Greg.Tinker@microchip.com).}}

\maketitle
	
\begin{abstract}
A  common source of defects in manufacturing miniature Printed Circuits Boards (PCB) is the attachment of silicon die or other wire bondable components on a Liquid Crystal Polymer (LCP) substrate. Typically, a conductive glue is dispensed prior to attachment with defects caused either by insufficient or excessive glue. The current practice in electronics industry is to examine the deposited glue by a human operator a process that is both time consuming and inefficient especially in preproduction runs where the error rate is high. In this paper we propose a system that automates fault diagnosis by accurately estimating the volume of glue deposits before and even after die attachment. To this end a modular scanning system is deployed that produces high resolution point clouds whereas the actual estimation of glue volume is performed by (R)egression-Net (RNet), a 3D Convolutional Neural Network (3DCNN). RNet outperforms other deep architectures and is able to estimate the volume either directly from the point cloud of a glue deposit or more interestingly after die attachment when only a small part of glue is visible around each die. The entire methodology is evaluated under operational conditions where the proposed system achieves accurate results without delaying the manufacturing process.
\end{abstract}


\markboth{IEEE TRANSACTIONS ON INDUSTRIAL ELECTRONICS}%
{}

\section{Introduction}

\IEEEPARstart{S}{mart} manufacturing has greatly benefited from the Internet of Things (IoT) revolution and advances in Deep Learning (DL) techniques as well. In this regard, several frameworks have been proposed that enabled the collections and analysis of diverse data from factory equipment and external sensors in order to make better decisions and optimize production towards zero-defect manufacturing strategies. 

This paper focuses on a characteristic use case falling into this category that often arise in the electronics manufacturing industry. In particular, a system is proposed for the scanning of PCBs and the inspection of die attachment on an LCP substrate. The key quantity that needs to be monitored is the volume of glue that is deposited on the LCP since insufficient or excessive glue is the primary source of malfunctions. The current industry practice for fault diagnosis is to have human operators inspect the manufactured PCBs and identify potential defects, a process that is error prone, cumbersome and time consuming causing delays in the production process.

\begin{figure}
\centering
\includegraphics[width=0.8\linewidth]{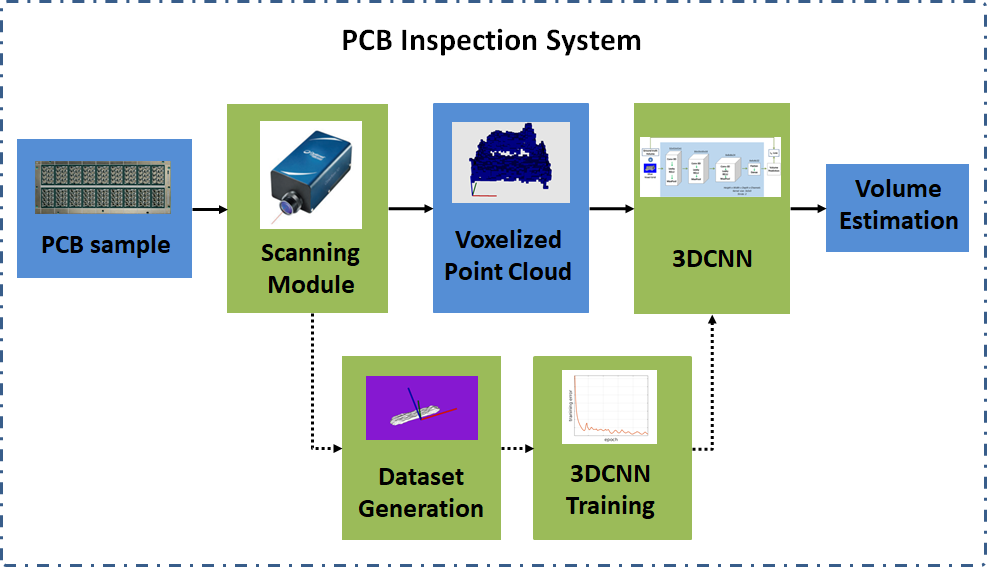}
\caption{The pipeline on the proposed system. Initially PCB's regions of interest are scanned in order to capture 3D information on the dispensed glue. The captured points clouds are fed to a 3DCNN that estimates the glue volume. Part of the proposed system is a procedure for generating a proper dataset for training the 3DCNN in a way that allows accurate volume estimation even when dies have been attached, occluding most of the dispensed glue.}
\label{fig:pipeline}
\end{figure}

The proposed system uses a custom scanning device consisting of a high-resolution laser scanning sensor and a modular motion control framework in order to scan the areas of interest on a PCB and to extract geometric information in terms of 3D point clouds. Using these point clouds, glue volumes are estimated using a 3DCNN where the captured point clouds are quantized into 3D voxels.  A graphical overview of our system is shown in Figure \ref{fig:pipeline}. In summary, the main technical contributions of our work can be pinpointed to,
\begin{itemize}
\item deploying a modular scanning system for inspecting PCBs
\item introducing a process for developing a volumetric dataset of glue deposits
\item proposing a 3DCNN architecture for estimating the volume of glue
\item correlating glue volume measurements before and after die attachment, enabling quality inspection after die attachment where only a trace of glue is visible.
\end{itemize}
The proposed approach counters difficulties commonly occurring in manual inspection tools stemming from subjective aspects and provides fast and accurate assessments. It is worth noting that while deep learning has certainly improved state of the art in fault diagnosis for industrial applications, to the best of our knowledge there has not been any deep learning architectures used for quality inspection on 3D data. In this paper, the use of (R)egression-Net is yet another step closer to the exploitation of deep learning techniques in 3D industrial applications. 

The remainder of the paper is organized as follows. Section \ref{sec:related} focuses on related work in deep learning and industrial applications, while Section \ref{sec:descrip} provides insight on the examined microelectronics use case. In Section \ref{sec:system} the proposed system is described in detail including the deployed scanning system, the generated volumetric dataset and the RNet architecture. Next, Section \ref{sec::experiments} is devoted to the experimental evaluation of our solution and finally Section \ref{sec:conclusions} draws the conclusions of our work.

\section{Related work}
\label{sec:related}

Deep Learning (DL) refers to a set of technologies, primarily Artificial Neural Networks (ANN), that has significantly improved state of the art in several tasks such as computer \cite{aake0} and robotic \cite{logh0} vision applications, image analysis \cite{liu0, chan0}, audio signals \cite{lee0,vafe0} and other complex signals \cite{ghami0}. In industrial systems, DL applications focus on inspection and fault diagnosis in industrial shopfloors and manufacturing processes. CNNs are the most representative category of deep learning models and their architectures are able to deal with big multimodal datasets. For instance, in \cite{wen0} a CNN framework is presented to convert sensor signals from manufacturing processes to 2-D images which are used to solve image classification tasks for fault diagnosis whereas in \cite{reyes0} a method is reported for analyzing tomography sensory data using a convolutional neural network pipeline. Moreover, in \cite{li0}, a deep convolutional computation model is proposed to learn hierarchical features of big data in IoT, by using the tensor-based representation model to extend the CNN from the vector space to the tensor space. Also, a convolutional discriminative feature learning method for induction motor fault diagnosis is introduced in \cite{sun0}, an approach that utilizes a combination of back-propagation neural network and a feed-forward convolutional pooling architecture. 

In another interesting application of DL in industry, the authors of \cite{yao0} propose a semisupervised deep learning model for data-driven soft sensor in order to monitor quality variables that cannot be measured directly. In a related line of research, \cite{yan0} presents a soft sensor modeling method that combines denoising autoencoders with a neural network. Furthermore, a deep learning framework is introduced in \cite{pan0} for adaptively extracting features from raw data that are used for fault classification while \cite{shi0} presents a  framework that utilizes multiple sparse autoencoders for tool condition monitoring. In a relevant work, the authors of \cite{luo0} propose a deep learning model for early fault detection of machine tools under time-varying conditions. The authors in \cite{he0}, have proposed an optimized deep architecture, namely a large memory storage retrieval neural network, to diagnose bearing faults. In a related application, \cite{shao1} propose a convolutional deep belief network  for bearing fault diagnosis applied in monitoring electric locomotive bearings while in the work of \cite{zhao0} a deep residual network is used with dynamically weighted wavelet coefficients for planetary gearbox fault diagnosis. An unsupervised feature engineering framework is proposed in \cite{oh0} that utilizes vibration imaging and deep learning for fault diagnosis in rotor systems. In another domain, \cite{chen0} reports a crack detection method for nuclear reactors monitoring using a CNN and a Naive Bayes data fusion approach. Also a deep neural network architecture for chemical fault diagnosis on imbalanced data streams is presented in \cite{hu0}.

Deep learning has been also deployed in the field of semiconductor manufacturing which is the application of our work. In \cite{naka0}, authors introduce a method for wafer map defect pattern classification and image retrieval using CNN, so as to identify root causes of die failures during the manufacturing process. A deep structure machine learning approach is also proposed in \cite{tell0} to identify and classify both single-defect and mixed-defect patterns by incorporating an information gain-based splitter as well as deep-structured machine learning. A fault detection and classification CNN model with automatic feature extraction and fault diagnosis functions has been proposed in \cite{lee1} to model the structure of the multivariate sensor signals from a semiconductor manufacturing process. 

While deep learning has certainly fostered progress in fault diagnosis of 1D and 2D signals it has not found wide application in processing 3D sensory from industrial shopfloors. Nonetheless, outside industry, 3DCNN models have recently gained traction  for traditional computer vision tasks. For instance, VoxNet \cite{matu0} integrates a volumetric occupancy grid representation with a supervised 3DCNN for object recognition. Similarly focussing on object recognition, a 3D volumetric object representation with a combination of multiple CNN models based on LeNet model is introduced in \cite{xu0} while in \cite{huan0}, the authors proposed an algorithm for labeling complex 3D point cloud data and provide a solution for training and testing 3DCNN using voxels. A CNN has been also proposed in \cite{pointnet} that directly consumes 3D point clouds and has shown impressive results in 3D object recognition. In another interesting work \cite{wang0} octree-based CNN is built upon the octree representation of 3D shapes and performs 3DCNN operations on the octants occupied by the 3D shape surface. Moreover, the authors of \cite{shao0} use a subset of ModelNet \cite{modelnet} and propose a CNN combining a spatial transformation in an attempt to deal with the rotation invariance problem in 3D object classification. In the following section we will see that 3DCNN based deep models have potential for fault diagnosis on 3D data as is validated through a concrete microelectronics use case, namely die attachment on a LCP substrate.

%there has been several approaches using other machine learning techniques for tasks such as object recognition using 3D point clouds \cite{camp0}. In addition recent progress in machine vision systems and scanning technologies provide the ability to have 3D point cloud data acquisition from objects of interest in real-time \cite{gome0}. Thus a common approach followed is the combination of scanning systems for extracting point clouds and machine learning techniques for further processing \cite{golo0,teic0}. For instance, the authors of \cite{habe0} proposed an approach based on a multi-layer perceptron for 3D object recognition. In this line of research deep learning architectures based on CNNs  have proven to be powerful classification tools for real world computer vision tasks. This is mainly attributed to difficulties in parameterizing 3D data in suitable forms for neural networks.
%
%In particular, the transformation of 3D point clouds into voxels in combination with 3DCNN architectures has gotten much attention in recent years. Indicatively,  In the following section we describe in detail the problem that we aim to tackle in the context of semiconductor industry and then proceed with the analysis of our system.

\section{Industrial background}

The proposed method is driven by a common problem faced in the electronics industry for the manufacturing of miniature electronic modules. The standard electronic equipment used in the industry for wire bonding, lacks the capability of automated optical inspection of the process. During manufacturing, silicon dies are attached to an LCP substrate using conductive glue that is placed by the glue dispenser machine. 

The manufacturing process consists of two main stages performed sequentially. The first stage is the glue dispensing stage and the second is the die (IC) attachment stage. The most critical variable for the attachment stage is the volume of the dispensed glue that is controlled indirectly by parameterizing the pressure on the glue dispenser machine. The two fault conditions that can be faced in the production are the excessive and insufficient glue volume. The excess in the amount of the glue leads to internal short circuits and on the other hand, insufficiency in the glue volume leads to weak die bonding. The current practice for the detection of these two fault conditions is the manual inspection of the glue volume at the first stage and the glue fillet, defined as the visible overflown glue around the die border, at the second stage. 

The manual inspection by a human operator who decides on the faulty conditions of the process, introduces an error that derives from the limitations of the human factor. This error corresponds to overhead costs on newly introduced parts in the production line that require adjustment period for the stabilization of the manual inspection process, resulting in an increase of the failure rate. The automation of the inspection process is considered an important step towards the automated quality control in the microelectronics industry and constitutes the motivation of the proposed system.

\section{Description of use case}
\label{sec:descrip}
\begin{figure}
\centering
\includegraphics[width=0.8\linewidth]{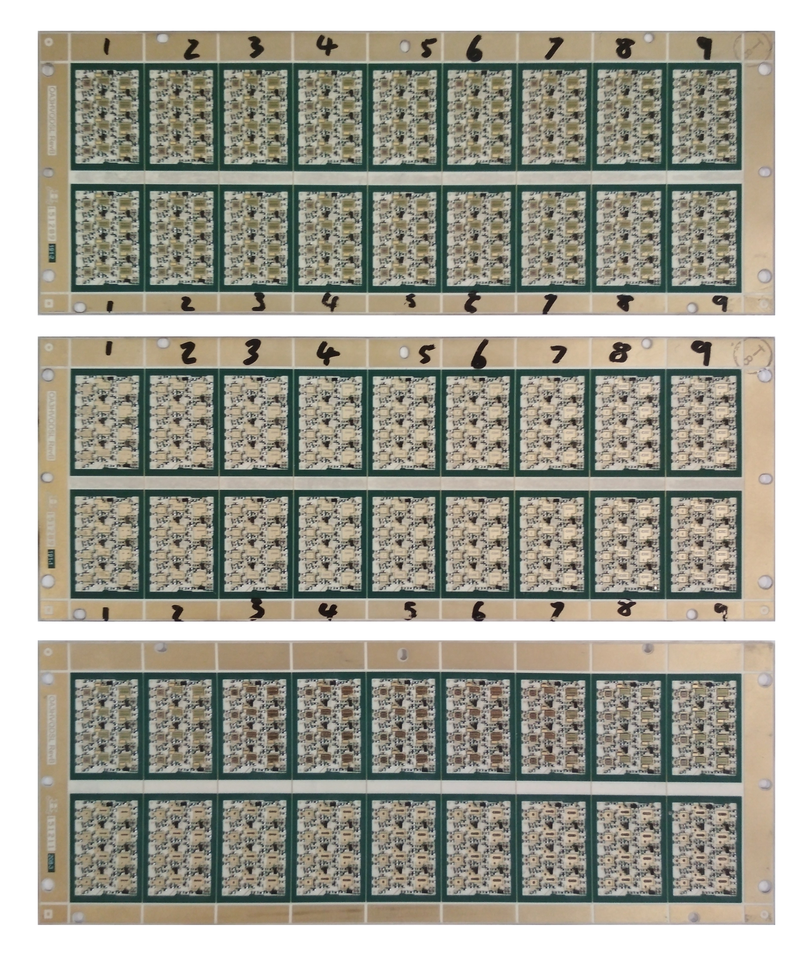}
\caption{Each of the PCBs used to validate the proposed system consists of two rows and nine columns for a total of $18$ circuit modules depicted in Figure 3. The PCBs are specifically manufactured to simulate the potential fault conditions during production. For the same column of circuit modules of the PCBs, the same glue quantity is dispensed. Moreover, from the left to the right columns, a larger quantity of glue has been progressively dispensed. The top and middle PCB consists of regions where the dies are attached and unattached respectively, while the bottom PCB consists of both attached (first row of nine circuits) and unattached (second row of nine circuits).}
\label{fig:pcb}
\end{figure}

\begin{figure}[h]
\includegraphics[width=0.6\linewidth]{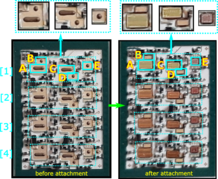}
\centering
\caption{On the left and right image, we have a closer look at one circuit module of a PCB with dimensions  $13mm\times19mm$, before and after die attachment. Notice the different types of glue annotated as \textit{A}, \textit{B}, \textit{C}, \textit{D} and \textit{E}. On each circuit there are four glue deposits on each type where approximately the same quantity of glue has been placed. As explained in Section \ref{sec::experiments} the top three deposits are used for training a 3DCNN and the bottom one for testing. On top, we see a magnified view of glue regions before and after the attachment of dies.}
\label{fig:one_panel}
\end{figure}

For the development and testing of the proposed system we have used three PCBs depicted in Figure \ref{fig:pcb}. As is illustrated in Figure \ref{fig:one_panel}, there are $5$ different types of dies annotated as A,B,C,D and E. The type of die defines the shape of the glue that needs to be placed on the LCP substrate for its attachment. This correspondence between the glue and die types is depicted in Figure \ref{fig:one_panel}. Each PCB comprises $18$ circuits over two rows and nine columns. Each of the $18$ circuits hosts four placeholders for each of the five glue types A,B,C,D and E, thus totally $20$ regions for the placement of the dies as illustrated in Figure \ref{fig:one_panel}.
     
To simulate the conditions that can be met during the production, the PCBs, in Figure \ref{fig:pcb}, have been specifically manufactured to have a wide range of glue quantities, from insufficient glue to excessive glue. Concretely, the glue quantity progressively increases starting from the far left column of each of the three PCBs and moving to the right. To ensure the same quantity per column, aside appropriately parameterizing the pressure on the glue dispenser, glue quantity has been visually verified during manufacturing. Moreover, to simulate both the glue dispensing stage and the die attachment stage, the top PCB comprises regions at the die attachment stage, the middle PCB comprises regions at the glue dispensing stage and the bottom PCB comprises regions at both stages; top $9$ circuits comprises regions after die attachment while the bottom $9$ circuits comprises regions before die attachment.

%For the development and testing of the proposed system we have used the PCBs depicted in Figure \ref{fig:pcb}. Each PCB comprises $18$ circuits over two rows and nine columns where dies have been attached on the top PCB and on the first row of the the bottom PCB.  As is illustrated in Figure \ref{fig:one_panel} each circuit is designed to host five different types of dies that correspond to different quantities and shapes of glue that are marked as \textit{A}, \textit{B}, \textit{C}, \textit{D} and \textit{E}. There are four installations of each glue type on a given circuit for a total of $20$ regions per circuit.
%
%%provided by Microsemi corporation \cite{microsemi}
%
%The PCBs have been specifically manufactured to have a wide range of glue quantities that can be met during production. Concretely, for the five different die regions on a single column of the PCBs, consisting of six circuits, an approximately constant quantity of glue has been placed by using the same pressure on the glue dispenser. This quantity progressively decreases starting from the far left column and moving to the right. For instance, the six circuits on the far left of the PCBs shown in Figure \ref{fig:pcb} have approximately the same quantity of glue placed on the respective regions of each type. To ensure this same quantity per column, aside appropriately parameterizing the pressure on the glue dispenser, glue quantity has been visually verified during manufacturing. 

\section{PCB inspection system}
\label{sec:system}
In this section, the components of the proposed system are described. Initially we present the custom scanning system that was developed for acquiring 3D measurements from a PCB. Next the processing of the collected data is described in order to generate an annotated data set of point cloud measurements and finally the proposed 3DCNN architecture is presented for glue volume estimation.

\subsection{Scanning system}
\label{sec::mprof}

\begin{figure}
\centering
\subfloat[]{\includegraphics[width=0.8\linewidth]{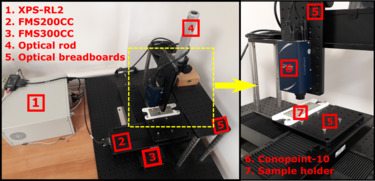}\label{fig:profilomphoto}}
\hfil
\subfloat[]{\includegraphics[width=0.8\linewidth]{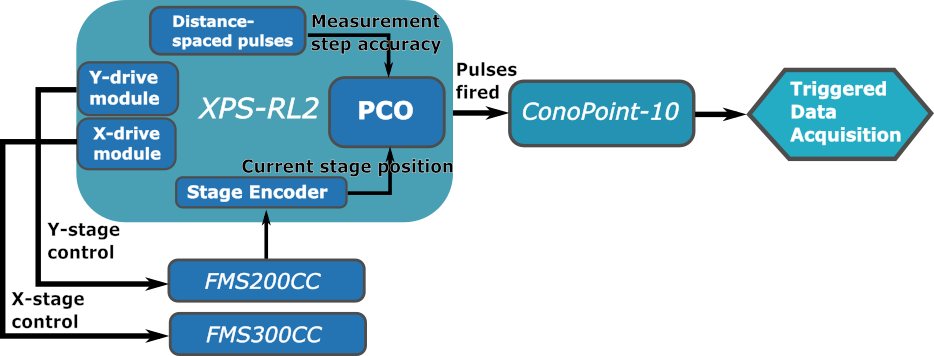}\label{fig:profilomdiag}}
\centering
\caption{\textbf{(a)} The left image is a snapshot of the scanning system where its basic components are highlighted, namely the motion controller (XPS-RL2), laser sensor (Conopoint-10), linear stages (FMS200CC and FMS300CC) and the parts (breadboards) for the mechanical integration of the system. The right image takes a closer look on the region in the yellow rectangle. \textbf{(b)} The operational flow of the scanning system where the motion controller drives the two linear stages X-stage and Y-stage using two drive modules. The Y-stage is configured as the scanning stage and its current stage position is used by the Position Compare Output (PCO) to fire the distanced-spaced pulses that trigger the data acquisition by the laser sensor.}
\label{fig:profilom}
\end{figure}

For PCB inspection, a modular laser scanning system has been built. As depicted in Figure \ref{fig:profilomphoto}, it includes an Optimet Conopoint-10 sensor \cite{optimet}, a Newport XPS-RL2 motion controller \cite{controller}, two linear stages and the support breadboards. In our system we use the FMS200CC and FMS300CC \cite{stages} linear stages with travel ranges of $200 mm$ and $300 mm$ respectively. 

Since inspection time is important for any production line, effort has been put in order to significantly decrease scanning time without losing accuracy. In this regard, a hardware triggering mechanism has been developed. The mechanism uses the Pulse Acquisition Mode of the sensor with measurement rate that is driven by distance spaced trigger pulses generated by the Position Compare Output (PCO) of the controller. The PCO provides $50 nsec$ delay between the crossing of the positions defined by the sample step and sending out the signal. The maximum speed of $100mm/s$ of the stages produces $5 nm$ uncertainty in position. The operational flow of the system is shown in \ref{fig:profilomdiag}.

\begin{figure}[h]
\centering
\includegraphics[width=0.8\linewidth]{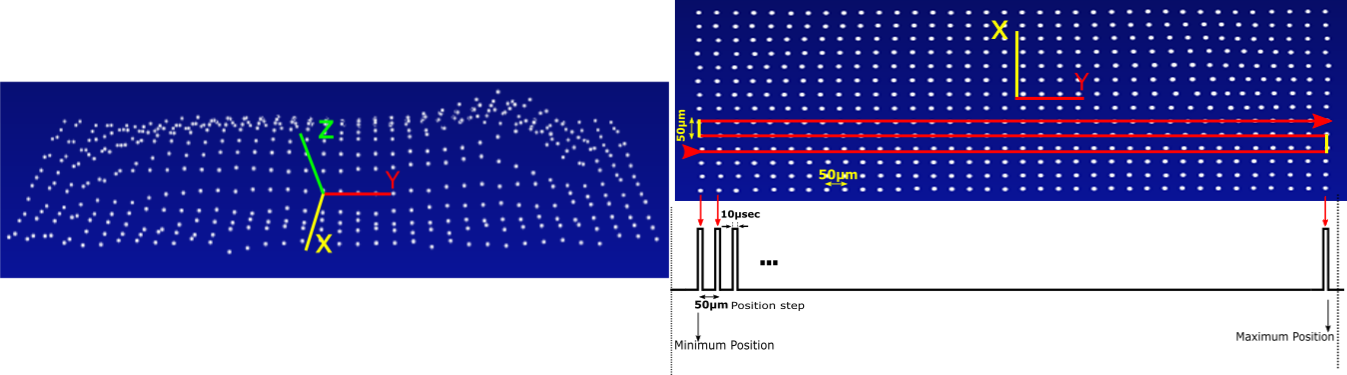}
%\subfloat[]{\includegraphics[width=0.8\linewidth]{./pulses_glue1.png}\label{fig:pulses1}}
%\hfil
%\subfloat[]{\includegraphics[width=0.8\linewidth]{./pulses_glue2.png}\label{fig:pulses2}}
%\hfil
%\subfloat[]{\includegraphics[width=0.8\linewidth]{./pulses_glue3.png}\label{fig:pulses3}}
\caption{The left pictures is a point cloud on $XYZ$ space for glue type A, generated by scanning a region of $0.7mm\times1.8mm$. On the top right image, the projection of the point cloud on $XY$ plane is shown. The scanning step of $50\mu m$ in $X$ and $Y$ axis and the zig-zag scanning direction are marked. The bottom right image shows the pulse diagram of the even distanced pulses generated by the controller that trigger  the sensor.}
\label{fig:pulses}
\end{figure}

In the bottom right image of Figure \ref{fig:pulses} we see the pulse diagram for a $1.8 mm$ range for Y-stage. The PCO generates a series of pulses that trigger the data acquisition at positions that are defined by the sample step of $50\mu m$. A total of 36 pulses are generated in Y axis. After the last pulse is generated, the X-stage moves by $50\mu m$ without being homed and a new series of pulses are generated in the reverse direction in Y-axis. Indicatively, by adapting this zig-zag scanning procedure as shown in the top right image of Figure \ref{fig:pulses} and implementing the hardware triggering mechanism, the scanning time is decreased approximately $10$-fold, compared to taking single measurements for every step along the motion. For all of the following experiments, this scanning strategy is followed.

\begin{figure}
\centering
\subfloat{\includegraphics[width=0.17\linewidth]{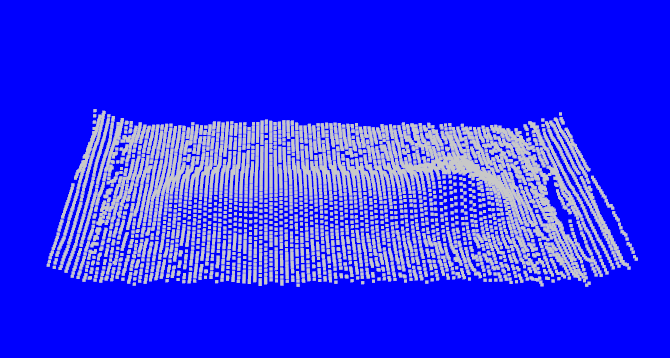}}
\hfil
\subfloat{\includegraphics[width=0.17\linewidth]{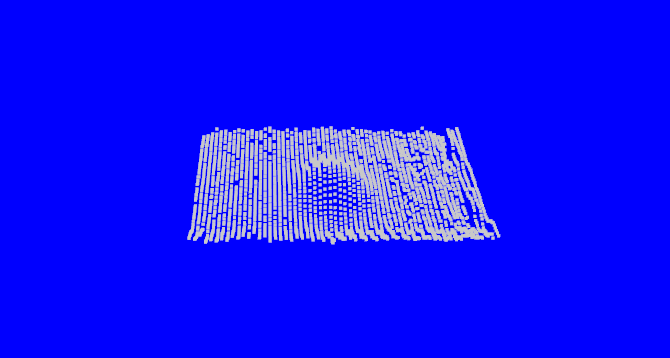}}
\hfil
\subfloat{\includegraphics[width=0.17\linewidth]{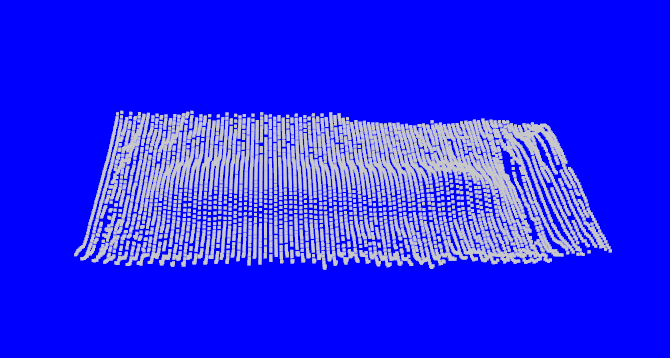}}
\hfil
\subfloat{\includegraphics[width=0.17\linewidth]{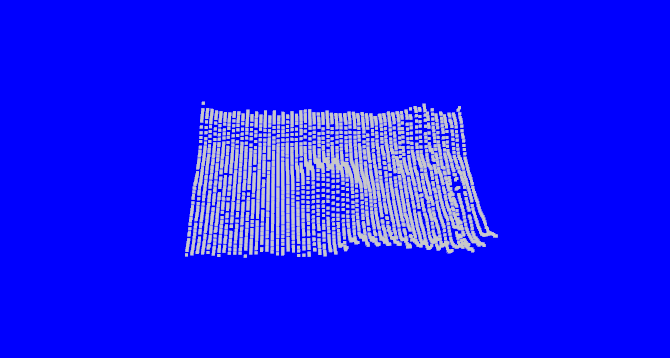}}
\hfil
\subfloat{\includegraphics[width=0.17\linewidth]{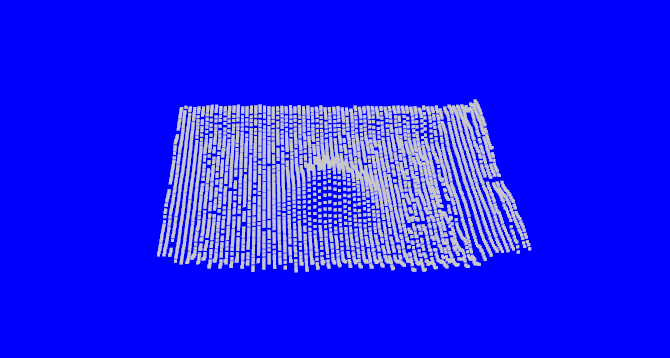}}
\hfil
\subfloat{\includegraphics[width=0.17\linewidth]{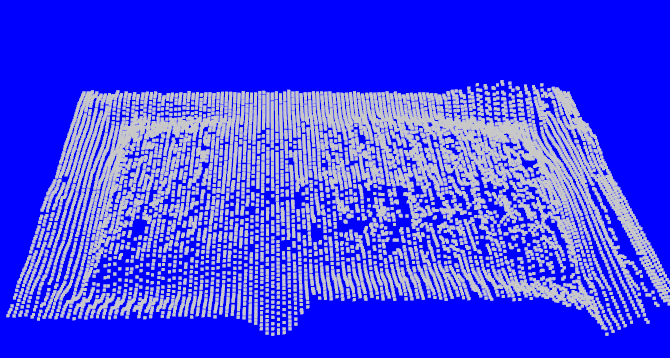}}
\hfil
\subfloat{\includegraphics[width=0.17\linewidth]{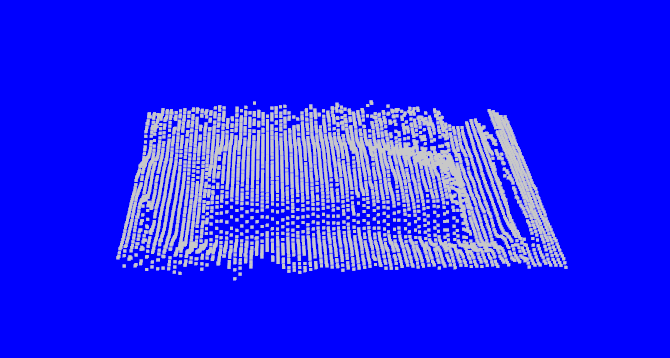}}
\hfil
\subfloat{\includegraphics[width=0.17\linewidth]{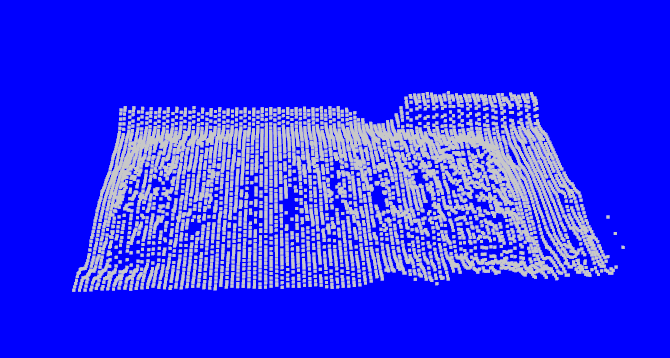}}
\hfil
\subfloat{\includegraphics[width=0.17\linewidth]{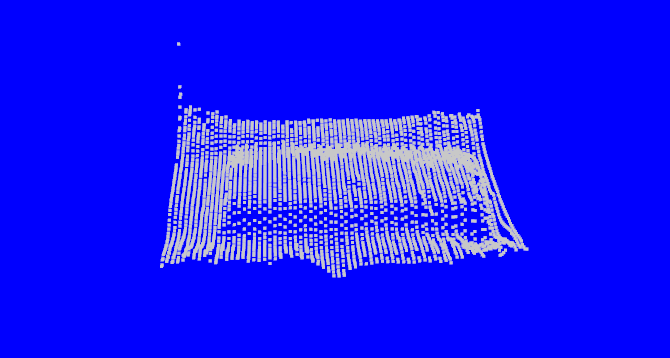}}
\hfil
\subfloat{\includegraphics[width=0.17\linewidth]{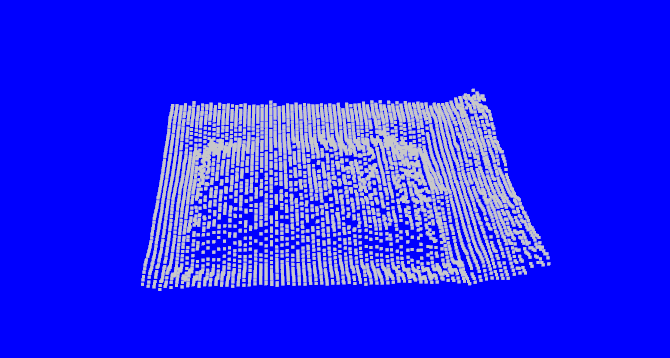}}
\caption{Scans over different regions of the PCB. From left to right, scans of type \textit{A}, \textit{B}, \textit{C}, \textit{D} and \textit{E} are shown both before (top) and after (bottom) dies are attached. Even though the dimension of these regions is below $1mm \times 2mm$ their 3D structure is well captured with a $20\mu m$ step.}
\label{fig:scans}
\end{figure}

\begin{figure}[h]
\includegraphics[width=0.6\linewidth]{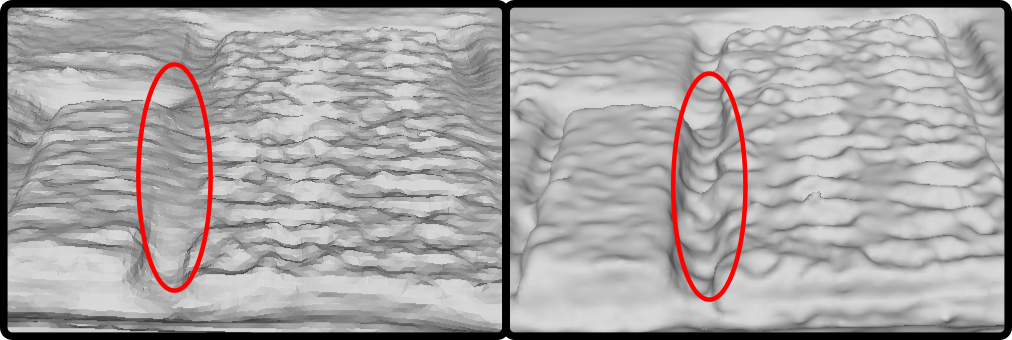}
\centering
\caption{Magnified views of the 3D surface after die attachment for same type regions where excessive glue is deposited (left) and insufficient glue is deposited (right). Notice the steeper gap between the die and the LCP substrate in the second case.}
\label{fig:meshfillet}
\end{figure}

Scans on a specific type of glue region have a constant range along the $X,Y$ axes. Some indicative scans over different regions of a PCB are shown in Figure \ref{fig:scans}. The 3D structure of the regions is captured quite well both before die attachment, where the glue deposit is clearly visible, and after die attachment. What is interesting for the more challenging second case, is that the glue fillet that exceeds the die border is captured as well as the 3D structure of the actual die. To verify this, in Figure \ref{fig:meshfillet} two same type regions are shown with the first one corresponding to a case of excessive glue on the left of the PCB and the second one to insufficient glue on the right of the PCB. Notice the much sharper gap between the die and the LCP substrate in the second case.

\subsection{Generating a dataset}
\label{sec::dataset}

\begin{figure}[h]
\centering
\subfloat[]{\includegraphics[width=0.3\linewidth]{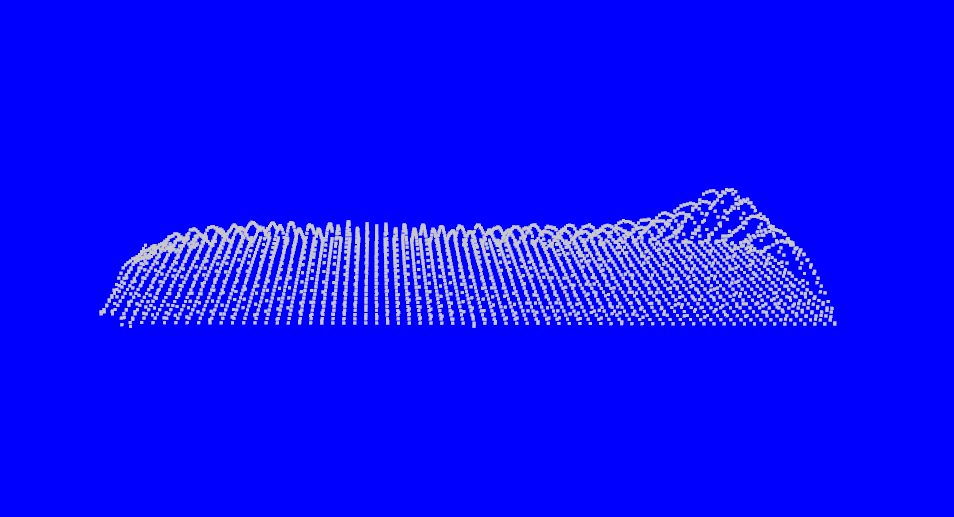}
\label{fig:annotsA}}
\hfil
\subfloat[]{\includegraphics[width=0.3\linewidth]{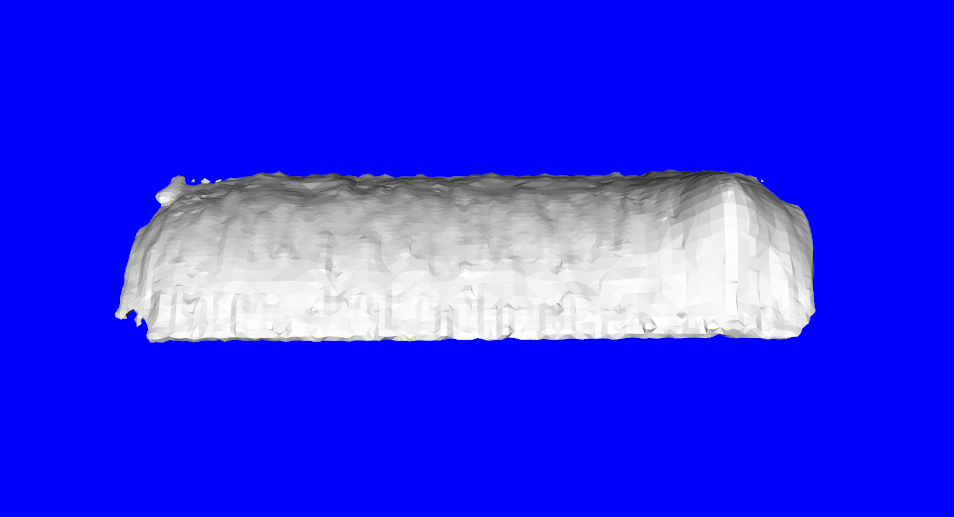}
\label{fig:annotsB}}
\subfloat[]{\includegraphics[width=0.35\linewidth]{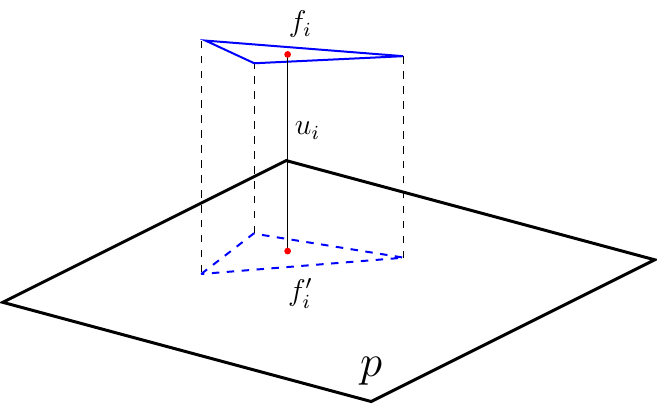}
\label{fig:projection}}
\caption{\textbf{(a)} The cropped point cloud of the regional scan depicted in the upper left image of Figure \ref{fig:scans} after its bottom is closed using the substrate plane. \textbf{(b)} The computed surface mesh after Poisson reconstruction is applied. \textbf{(c)} Denoting a random face of glue mesh as $f_i$, its projection on the LCP substrate plane $p$ is represented as $f_i^{\prime}$ whereas $u_i$ the distance between the centers of  $f_i,f_i^{\prime}$.}
\label{fig:annots}
\end{figure}

To train and evaluate our monitoring framework an annotated point cloud dataset has been developed where the volumes of glue deposits have been analytically computed. To this end we used the circuits on the second PCB and second row of the bottom PCB in Figure \ref{fig:pcb} where no dies are attached and glue is clearly visible and assumed that the volume of deposited glue is equal for regions in the same column as explained in Section \ref{sec:descrip}. For each regional scan, the plane equation for the LCP substrate has been calculated using the RANSAC algorithm \cite{fisc0} for plane fitting as implemented in PCL \cite{rusu0} and the 3D points corresponding to the glue mass have been manually cropped. In order to get a closed 3D surface, each 3D point of the cropped point cloud is projected on the estimated substrate plane as is illustrated on the example of Figure \ref{fig:annotsA}. Having a closed 3D point cloud, Poisson reconstruction \cite{kazh0} is applied to compute the surface mesh of glue as is depicted in Figure \ref{fig:annotsB}. Geometrically, having a surface $S$, its volume $V$ can be computed using the formula,

\begin{equation}
V = \iiint\limits_S \, \mathrm{d} x\,\mathrm{d} y\,\mathrm{d} z.
\end{equation}

In our case an analytical representation of $S$ is not available, therefore glue volume is approximated using its reconstructed mesh. More formally, let us denote the glue mesh as a set $F = \{f_1,f_2,\dots,f_N\}$ of 3D points' triangles and the substrate plane as $p$. Additionally, we denote with $f_i^{\prime}$ the projection of $f_i$ on $p$ and with $u_i$ the distance between the centers of $f_i$ and $f_i^{\prime}$ as is more clearly depicted in Figure \ref{fig:projection}. The glue volume is approximated as,
\begin{equation}
V = \sum_i |f_i^{\prime}| u_i
\end{equation}
where $|f_i^{\prime}|$ is the area of the projected rectangle. Following this procedure, we end up with $108$ annotated point clouds for each of the five different types of glue deposits on the circuits with no dies attached. We extend these volume estimations to the rest of the circuits where dies have been attached in order to annotate the entire PCB dataset. Naturally, this extension process as well as the volume approximation is error prone due to several factors such as sensor noise, discrepancies between glue deposits due to the glue dispenser and the induced approximation error. Nonetheless, as shown in Section \ref{sec::experiments} these estimations captures the manufacturer's trend of progressively depositing more glue. This allows the correlation of the hidden glue after die attachment with actual measurements, thus enabling fault diagnosis even after dies are attached.

\begin{table}
\small
\centering
\begin{tabular}{|c||c|c|c|c|c|}
\hline
\textit{Glue Type} & \textit{A} & \textit{B} & \textit{C} & \textit{D} & {E}  \\
\hline
\hline
\multicolumn{6}{|c|}{\textbf{20 micrometer sampling step}} \\
\hline
\hline 
\textit{unatt. (train)} & 40500 & 25920 & 40500 & 25920 & 40500 \\
\hline
\textit{unatt. (test)} & 13500 & 8640 & 13500 & 8640 & 13500 \\
\hline
\hline
\textit{att. (train)} & 40500 & 32400 & 40500 & 32400 & 40500 \\
\hline
\textit{att. (test)} & 13500 & 10800 & 13500 & 10800 & 13500 \\
\hline
\hline
\multicolumn{6}{|c|}{\textbf{50 micrometer sampling step}} \\
\hline
\hline
\textit{unatt. (train)} & 12960 & 6480 & 12960 & 6480 &  6480\\
\hline
\textit{unatt. (test)} & 4320 & 2160 & 4320 & 2160 &  2160\\
\hline
\hline
\textit{att. (train)} & 19440 & 9720 & 19440 & 9720 & 14580 \\
\hline
\textit{att. (test)} & 6480 & 3240 & 6480 & 3240 & 4860\\
\hline
\end{tabular}
\caption{Number of training and test samples.}
\label{tab::dataset}
\end{table}

\begin{figure}
\centering
\subfloat[]{\includegraphics[width=0.5\linewidth]{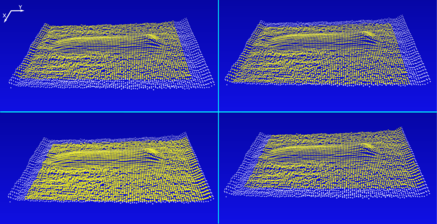}
\label{fig:dataaug}}
\subfloat[]{\includegraphics[width=0.41\linewidth]{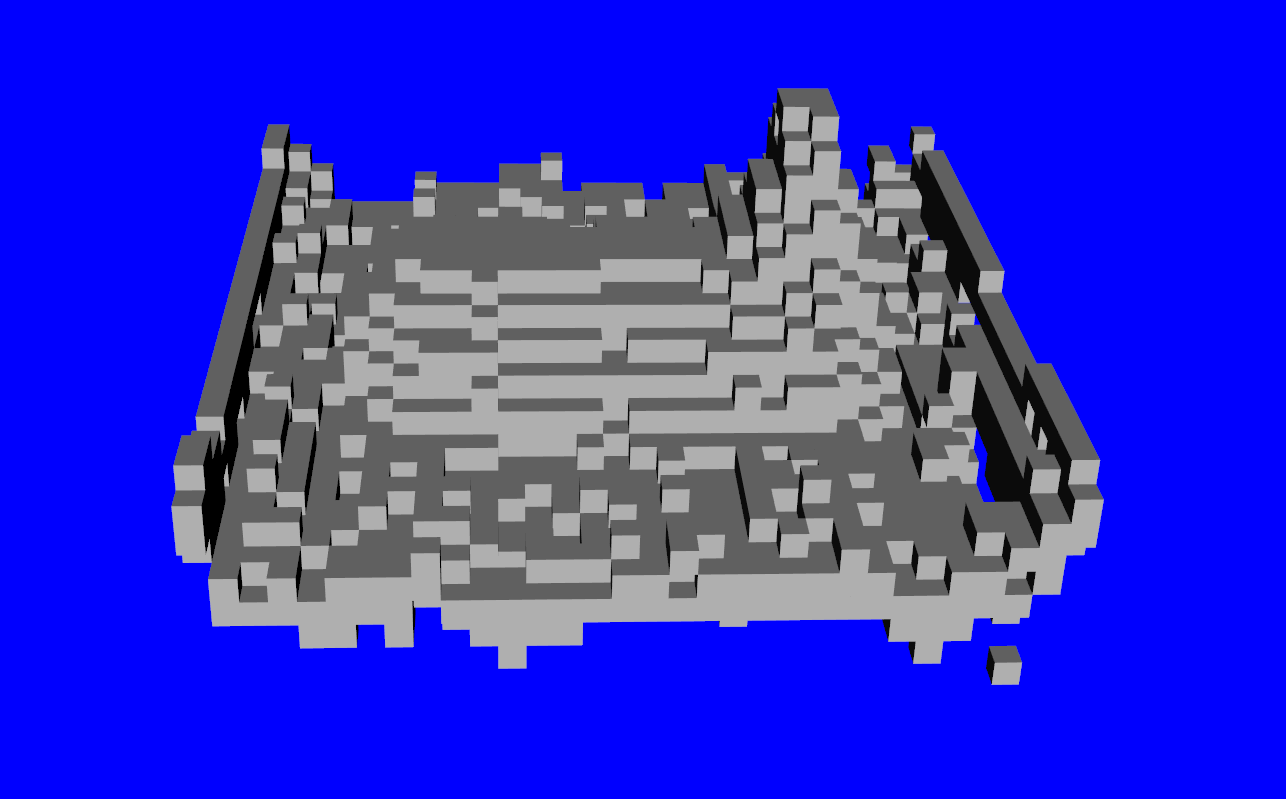}
\label{fig:occ_grid}}
\caption{\textbf{(a)} Augmentation example. In yellow we see the point cloud cropped from the original one at various augmentation steps. \textbf{(b)} The $32 \times 32 \times 32$ voxel grid of the regional scan depicted in the upper left image of Figure \ref{fig:scans}. The quantization of the 3D space along the $Z$ axis is purposely more detailed since information along the vertical axis is more important for estimating the glue volume. Note for instance the magnified glue bump on the right of the glue deposit that occurred when glue dispensation finished.}
\end{figure}

Of course the number of annotated samples is very small for a 3DCNN network, therefore a data augmentation step has been implemented. To this end and in order to also take into account measurement noise we scan each PCB five times, thus artificially raising the number of PCB samples to fifteen. Moreover, the range along the $X,Y,Z$ is calculated for each regional scan. A cropping filter is applied on the $X-Y$ plane using a bounding box that is shifted along the $X,Y$ axes. For all our experiments the bounding box covers $92\%$ of the range on the $X,Y$ axes and is shifted by a constant step $s$,
\begin{equation}
s=\max(0.02*r, \epsilon)
\end{equation}
where $r$ is the respective axis range and $\epsilon$ is the minimum step of the linear stage which is either $20{\mu}m$ or $50{\mu}m$ for our experiments. This augmentation procedure, as is also shown in the example of Figures \ref{fig:dataaug}, simulates the potential mis-localization of the scanning device. The dataset is further augmented by adding Gaussian noise along the $Z$ axis coarsely approximating noise during the scanning procedure. In our experiments, $4$ noise levels have been used by increasing the noise deviation by $3\%$ with respect to the range on $Z$. Following this procedure for different step of the linear stages, namely $20{\mu}m$ and $50{\mu}m$, we get several thousands of point clouds for each of the different region types as is summarized in Table \ref{tab::dataset}. In the table we denote with att. and unatt. the cases with and without die attached and we adopt a $3$ to $1$ ratio for splitting between training and test sets.

\subsection{3DCNN for glue volume estimation}

\begin{figure}
\centering
\includegraphics[width=0.95\linewidth]{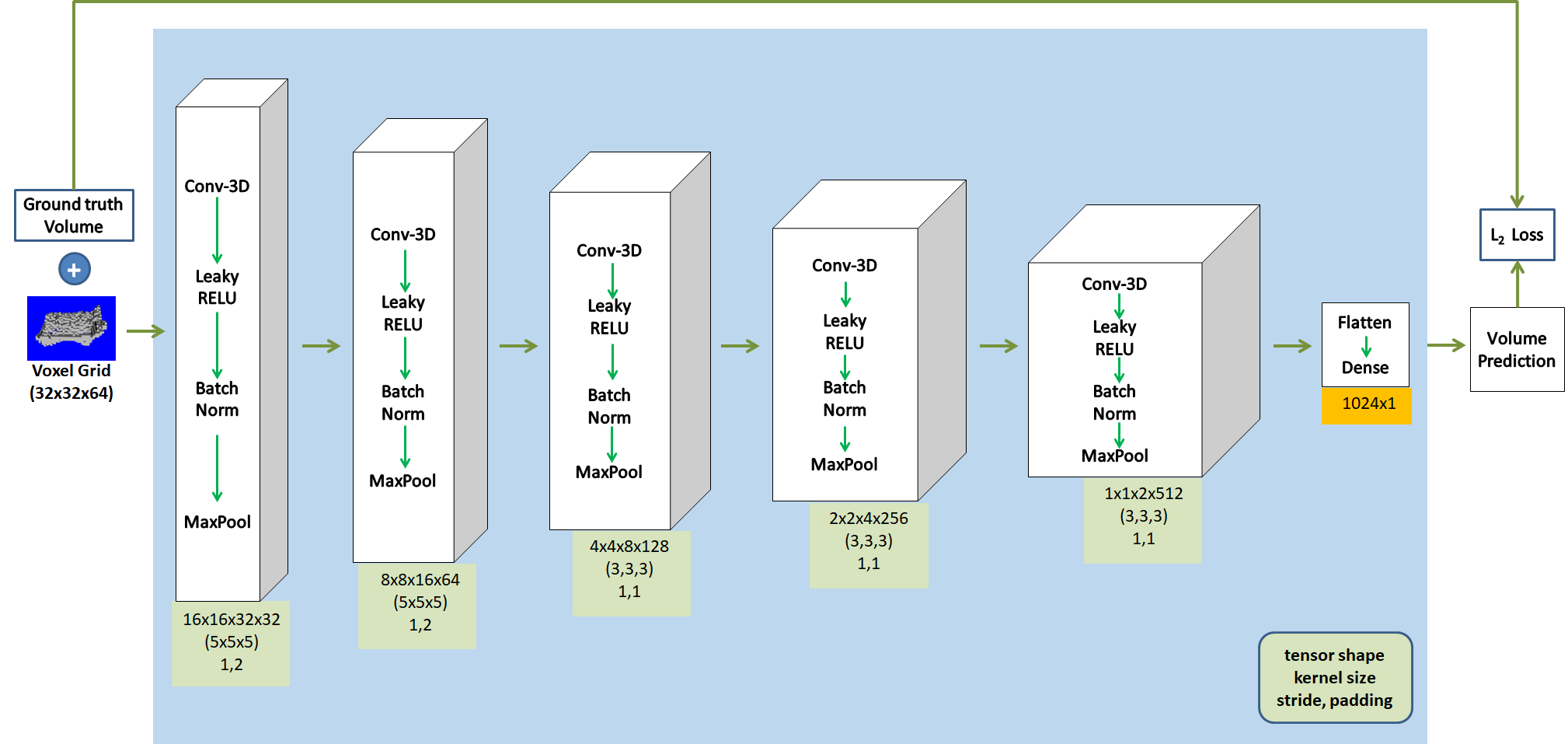}
\caption{The RNet architecture. Its input is a voxel grid that is sequentially passed through convolutional blocks of 3D convolution, leaky RELU, batch normalization and max pooling layers. To estimate glue volume a flattening and a dense layer are applied on the last convolutional block. Under each block we report the size of the output tensor, the kernel size of the convolutional filter as well as the stride and padding parameters.}
\label{fig:cnn-arch}
\end{figure}

To automate glue volume estimation we train $10$ instances of the proposed RNet architecture with each one corresponding to a particular type of  glue before or after the respective die has been attached. Point clouds are parameterized using $0-1$ voxel grids similar to the works of \cite{matu0, zhwu0}. This simple representation works fine in our case since, in contrast to depth cameras, there are no occlusion issues caused by a single viewpoint while point cloud density is approximately constant. Throughout our experiments the size of voxel grids is $32 \times 32 \times 64$. The dimensions of a single voxel are constant for each type of glue. Specifically, along the $X,Y$ axes the dimensions are equal to the range of point cloud divided by the number of voxels whereas it is set to $10{\mu}m$ along the $Z$ axis. For glue types where the dimension along the $Z$ is not adequate for covering at least $98\%$ of the captured 3D points, the dimension of voxels increases in steps of $5{\mu}m$ until this criterion is fulfilled. An indicative voxel grid is shown in Figure \ref{fig:occ_grid} corresponding to the point cloud of the upper left image in Figure \ref{fig:scans}. To facilitate visualization we have reduced the number of voxels along $Z$ to $32$.

The architecture of RNet is shown in Figure \ref{fig:cnn-arch} along with the specific parameters of each layer. The network's input is an occupancy grid and there are five network blocks consisting of sequences of 3D convolutional, leaky ReLU, batch normalization and max pooling layers that progressively drop the spatial dimensions of the input tensor while increasing the number of channels. Concretely the dimension of the input tensor for the first block is  $32 \times 32 \times 64 \times 1$ and the output tensor is $16 \times 16 \times 32 \times 32$ which is then propagated to the following block. After these five network blocks, the last one generates a $1 \times 1 \times 2 \times 512$ tensor that is subsequently flattened to a $1024$ feature vector which is forwarded to a $1024 \times 1$  dense layer that produces the final volume estimation. 
Figure \ref{fig:cnn-arch} includes the dimension of the output tensor under each network block along with the kernel size, stride and padding of the respective convolutional layer. The size of the max pooling window is $2 \times 2 \times 2$ with a stride of $2$ for all network blocks. It should be mentioned that we have also experimented with deeper architectures without noticing substantial accuracy improvement. The cost function that is minimized is the $l_2$ error between ground truth and estimation. Namely, if the glue volume of grid $g$ is denoted as $V(g)$ and the volume estimation as $V(g)^{\prime}$ the network loss is defined as,
\begin{equation}
L = \mathbb{E}[|V(g)-V(g)^{\prime}|^2].
\end{equation}
The underlying optimization problem is solved using Adam optimization algorithm \cite{adam0} as is implemented in the PyTorch framework \cite{pytorch}.

\section{Experiments}
\label{sec::experiments}
This section presents a series of experiments that were carried out in order to examine the performance of our framework in different aspects and compare it with alternative state of the art methods. 

\subsection{Experimental setup}
\label{sec::expsetup}

The complexity of the PCB with its different types of glue regions and dies provide a challenging evaluation setup for the proposed framework. The conducted experiments have been divided in two categories based on the dataset described in Section  \ref{sec::dataset}. In the first category dies have not been attached and glue deposits are clearly visible while in the second one dies have been placed and only glue on the sides of dies is partially visible. Another important experimental parameter that is  examined is inspection time using different scanning step. In this regard, two different sets of experiments are presented where the sampling step for the scanning system of Section \ref{sec::mprof} is $20$ and $50$ micrometers.

For each of these sets of experiments, $10$ separate instances of RNet are trained  corresponding to the five different glue types before and after die attachment. For each instance, point clouds are split in training and test set under a $75\% - 25\%$ ratio. To avoid biasing the evaluation by the data augmentation described in Section  \ref{sec::dataset}, the training and validation sets correspond to scans from different regions on the PCBs. Concretely, as is also explained in Figure \ref{fig:one_panel}, on each circuit module the top three glue deposits of each type are used for training and the last one for testing. 

Each RNet instance has been trained for $100$ epochs using a batch size of $128$, a learning rate of $0.0001$ and having $4908865$ trainable parameters. Training and validation mean square error remained approximately constant after $80$ epochs thus we assume that the training procedure converges to a stable minimum in all cases. Training time was up to $15$ hours for the cases with the largest training set and was performed on two Tesla K40m GPUs. For parameter initialization all convolutional layers are initialized using a Gaussian distribution of mean 0 and standard deviation of 0.02 while batch normalization layers are initialized using a mean of 1 and standard deviation of 0.02.

\subsection{Data augmentation analysis}
\label{sec::expaug}

\begin{table}
\small
\centering
\begin{tabular}{|c|c||c|c|c|c|c|}
\hline
\multicolumn{2}{|c|}{\textit{Type}} & \textit{A} & \textit{B} & \textit{C} & \textit{D} & \textit{E}  \\
\hline
\hline
\multicolumn{7}{|c|}{\textbf{20 micrometer sampling step}} \\
\hline 
\multirow{2}{*}{\textit{no IC}} & \textit{aug.} & 8.21 & 0.45 & 5.37 & 0.39 & 0.21 \\
\cline{2-7}
 & \textit{no aug.} & 88.52 & 1.06 & 45.02 & 0.78 & 2.24 \\
\hline 
\multirow{2}{*}{\textit{IC}} & \textit{aug.} & 21.54 & 1.25 & 25.19 & 1.02 & 1.37 \\
\cline{2-7}
 & \textit{no aug.} & 68.52 & 2.07 & 79.93 & 1.91 & 5.26 \\ 
 
\hline
\multicolumn{7}{|c|}{\textbf{50 micrometer sampling step}} \\
\hline 
\multirow{2}{*}{\textit{no IC}} & \textit{aug.} & 7.99 & 0.35 & 5.29 & 0.47 & 0.25 \\
\cline{2-7}
 & \textit{no aug.} & 33.91 & 0.98 & 19.05 & 0.90  & 1.21  \\
 \hline 
\multirow{2}{*}{\textit{IC}} & \textit{aug.} & 26.89 & 1.00 & 28.40 & 1.00 & 1.55 \\
\cline{2-7}
 & \textit{no aug.} & 99.16  & 2.27 & 80.56  & 2.05  & 6.47  \\ 

\hline
\end{tabular}
\caption{Validation MSE in $e{-06}$ $(mm^3)^2$ with(out) augmentation.}
\label{tab::aug}
\end{table}

In this paragraph we examine the contribution of the data augmentation step. In this regard, Table \ref{tab::aug} contains the achieved Mean Square Error (MSE) on the validation set when training with and without data augmentation. The rows denoted as \textit{no IC} and \textit{IC} correspond to before and after die attachment. In all cases it is apparent that the data augmentation step on Section \ref{sec::dataset} clearly improves the generalization of the trained model, achieving lower MSE by at least an order of magnitude.

\subsection{Volume estimation evaluation}
\label{sec::exreg}

\begin{figure*}[ht]
\centering
\subfloat{\includegraphics[width=0.2\linewidth]{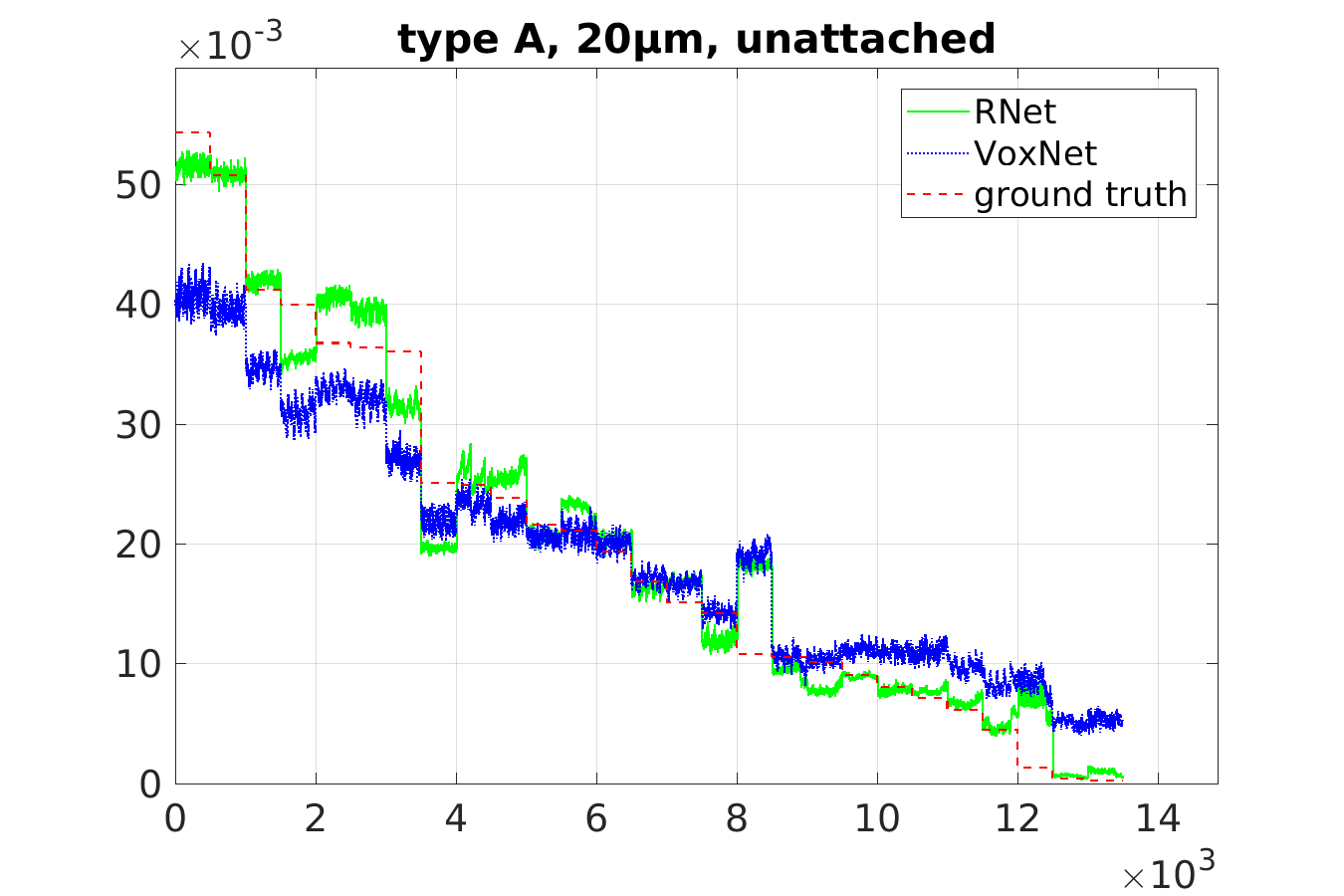}}
\hfil
\subfloat{\includegraphics[width=0.2\linewidth]{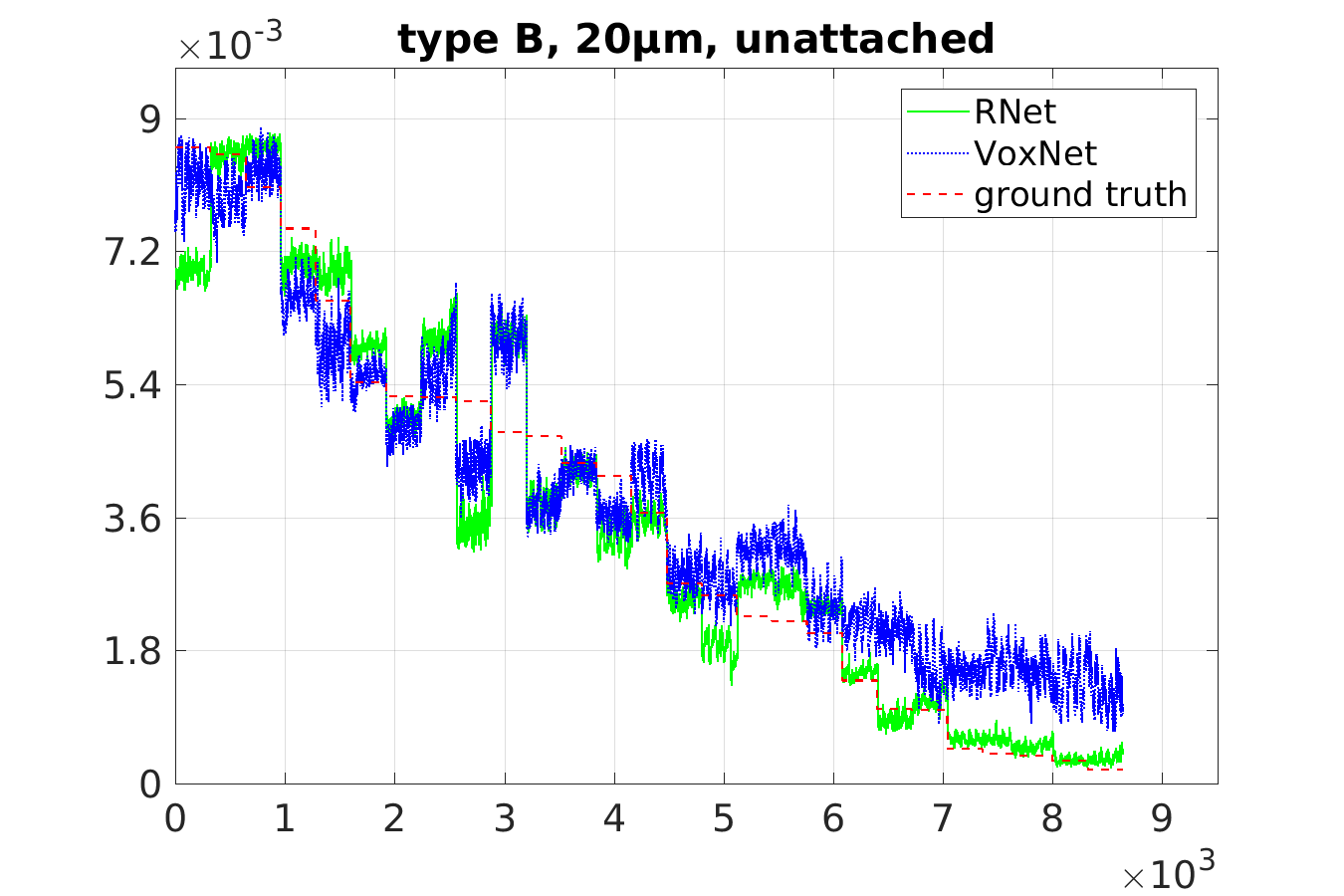}}
\hfil
\subfloat{\includegraphics[width=0.2\linewidth]{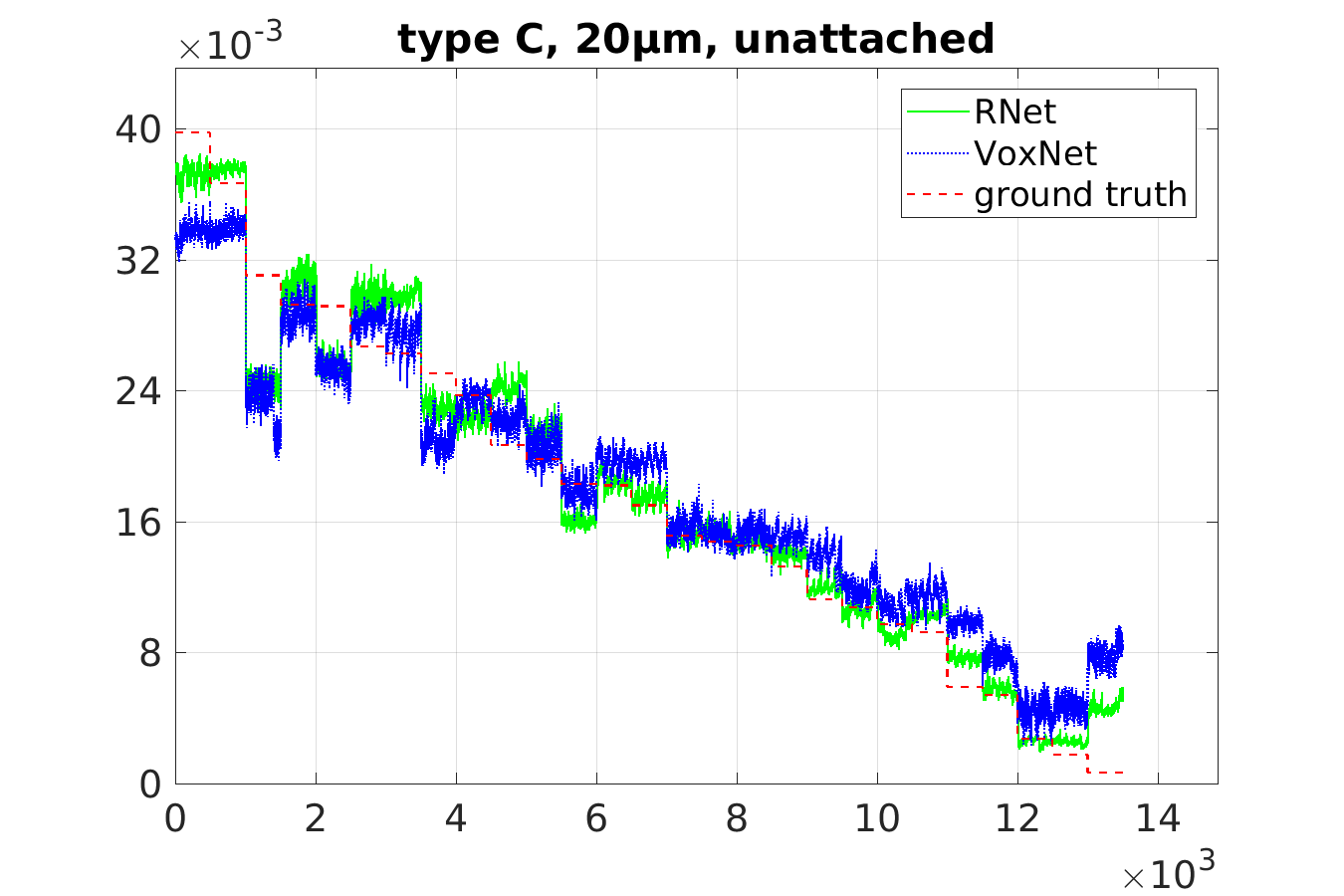}}
\hfil
\subfloat{\includegraphics[width=0.2\linewidth]{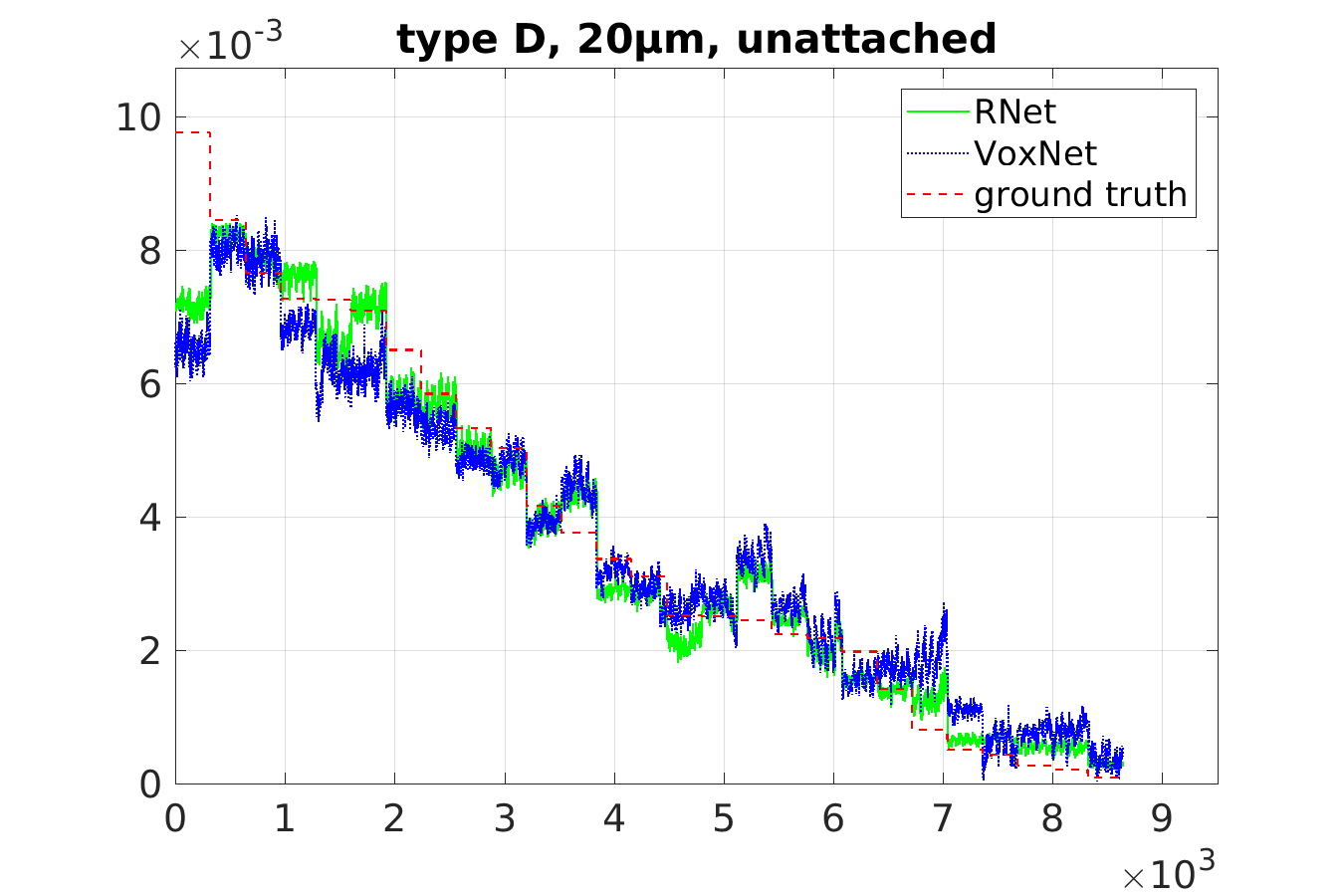}}
\hfil
\subfloat{\includegraphics[width=0.2\linewidth]{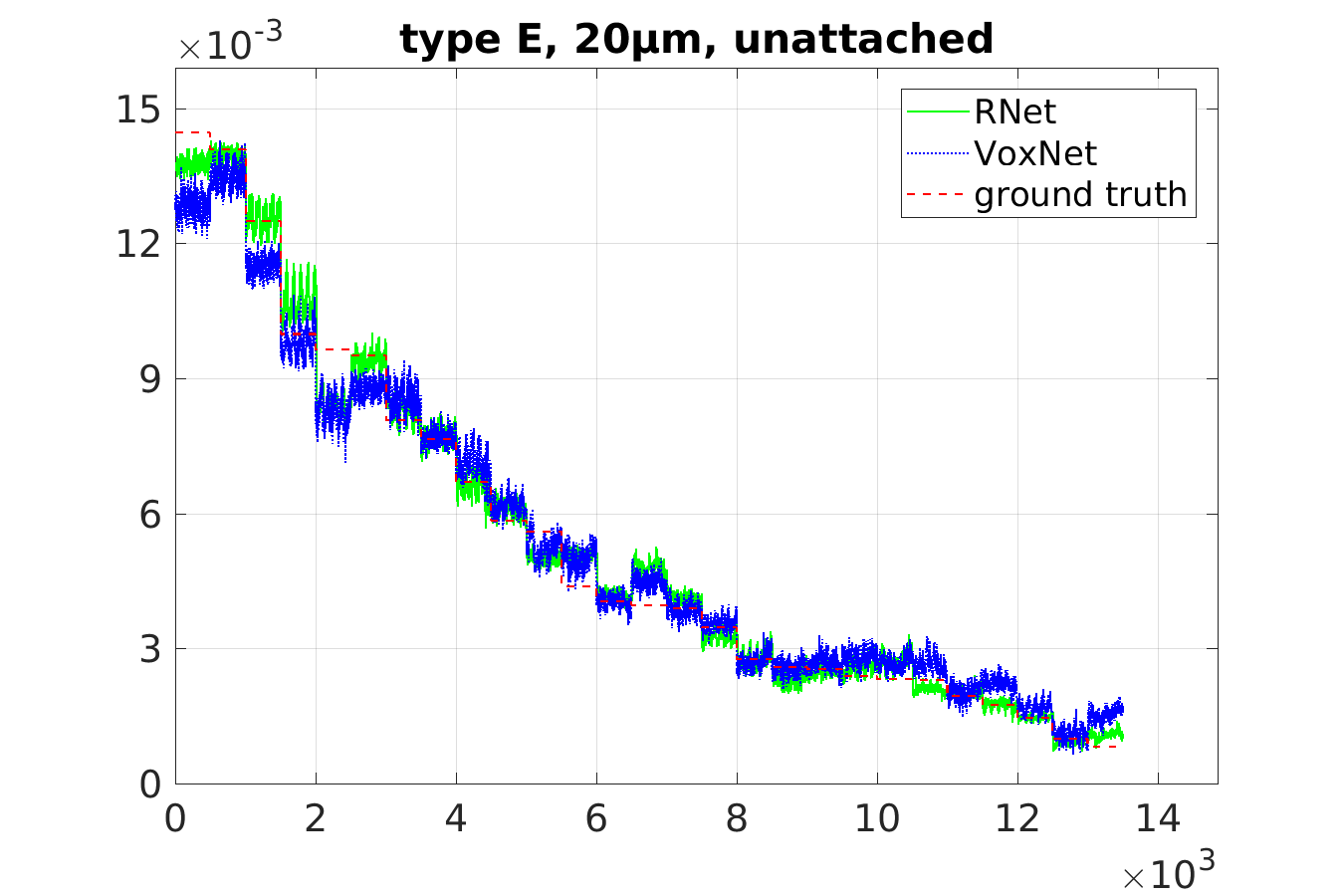}}
\hfil
\subfloat{\includegraphics[width=0.2\linewidth]{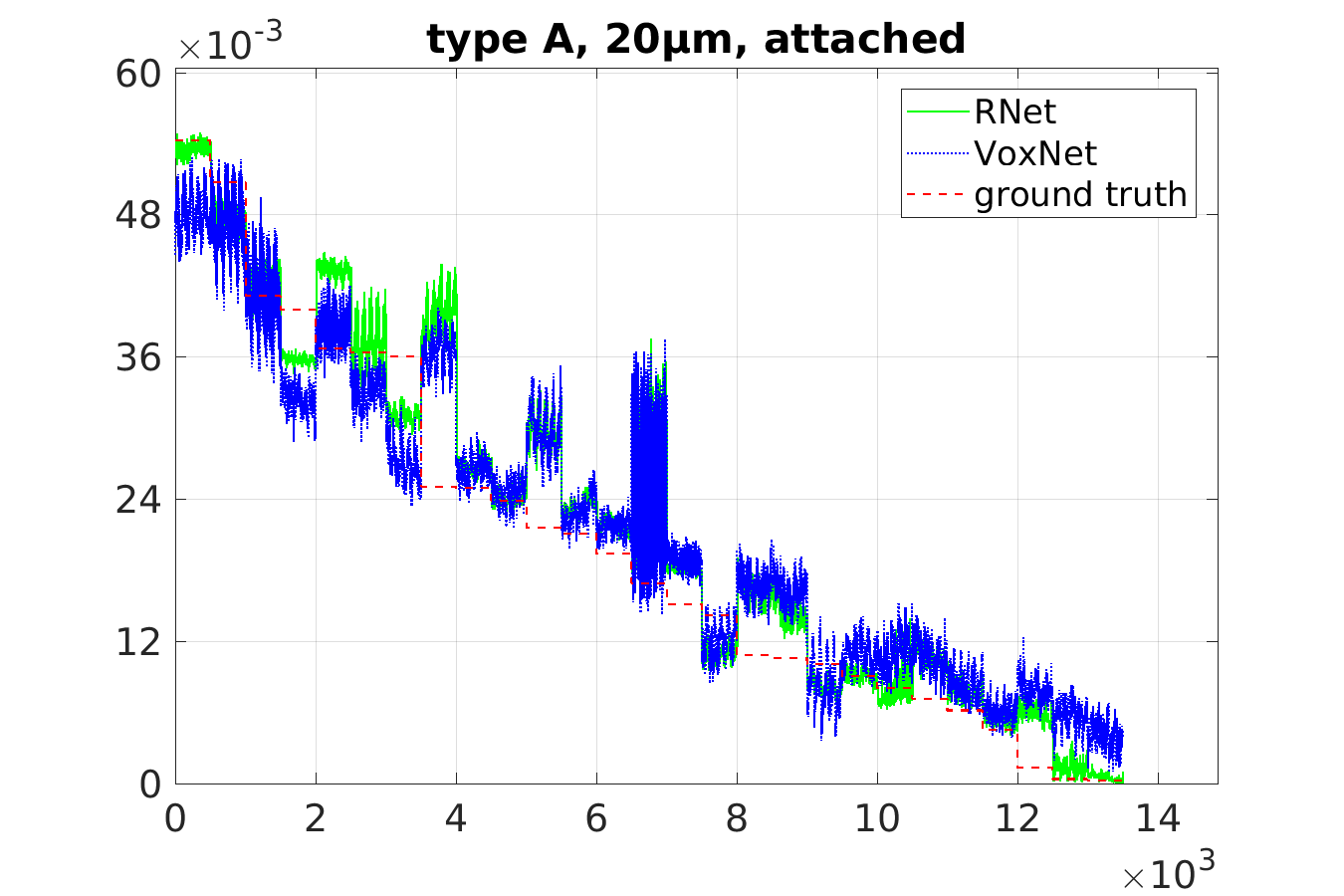}}
\hfil
\subfloat{\includegraphics[width=0.2\linewidth]{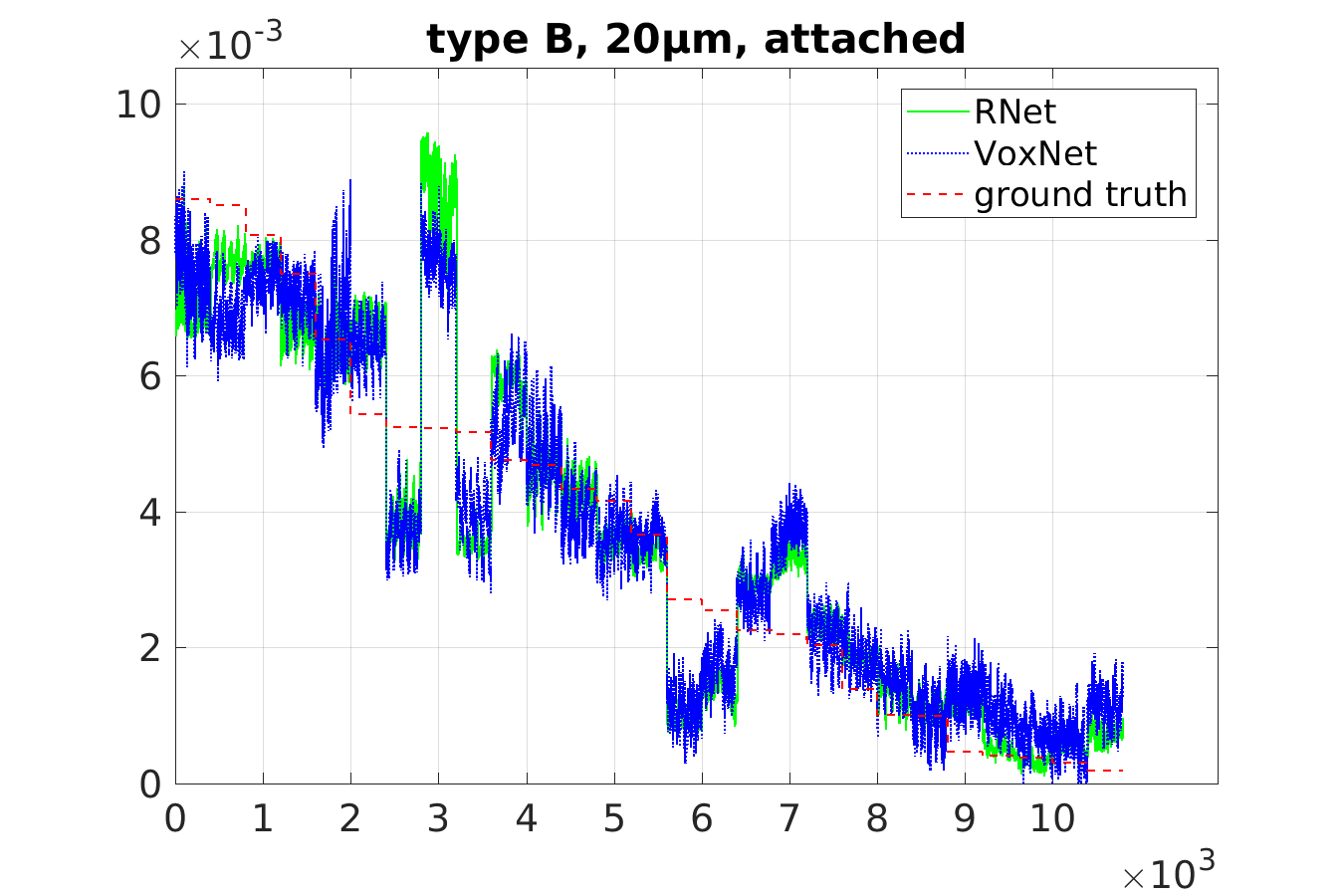}}
\hfil
\subfloat{\includegraphics[width=0.2\linewidth]{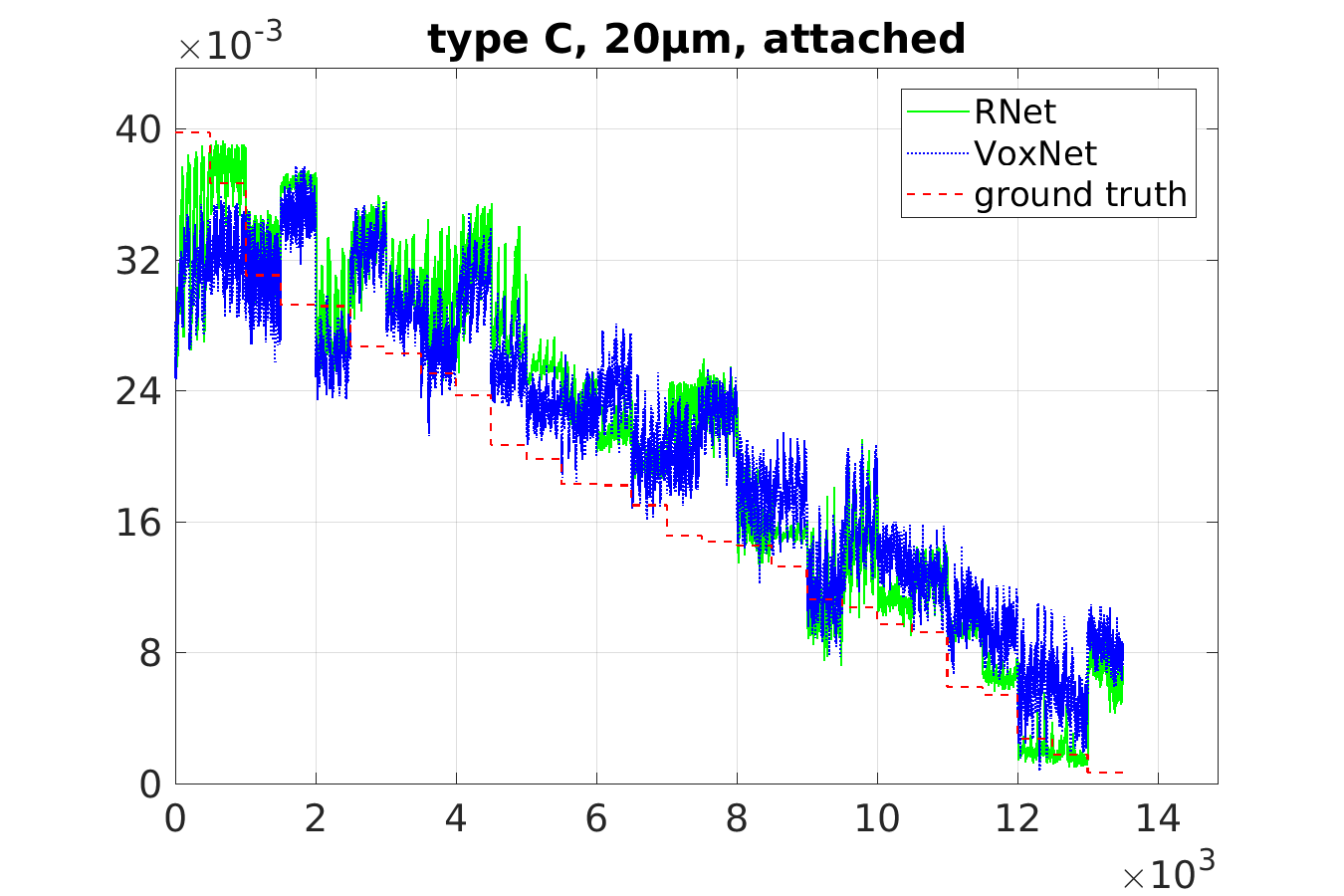}}
\hfil
\subfloat{\includegraphics[width=0.2\linewidth]{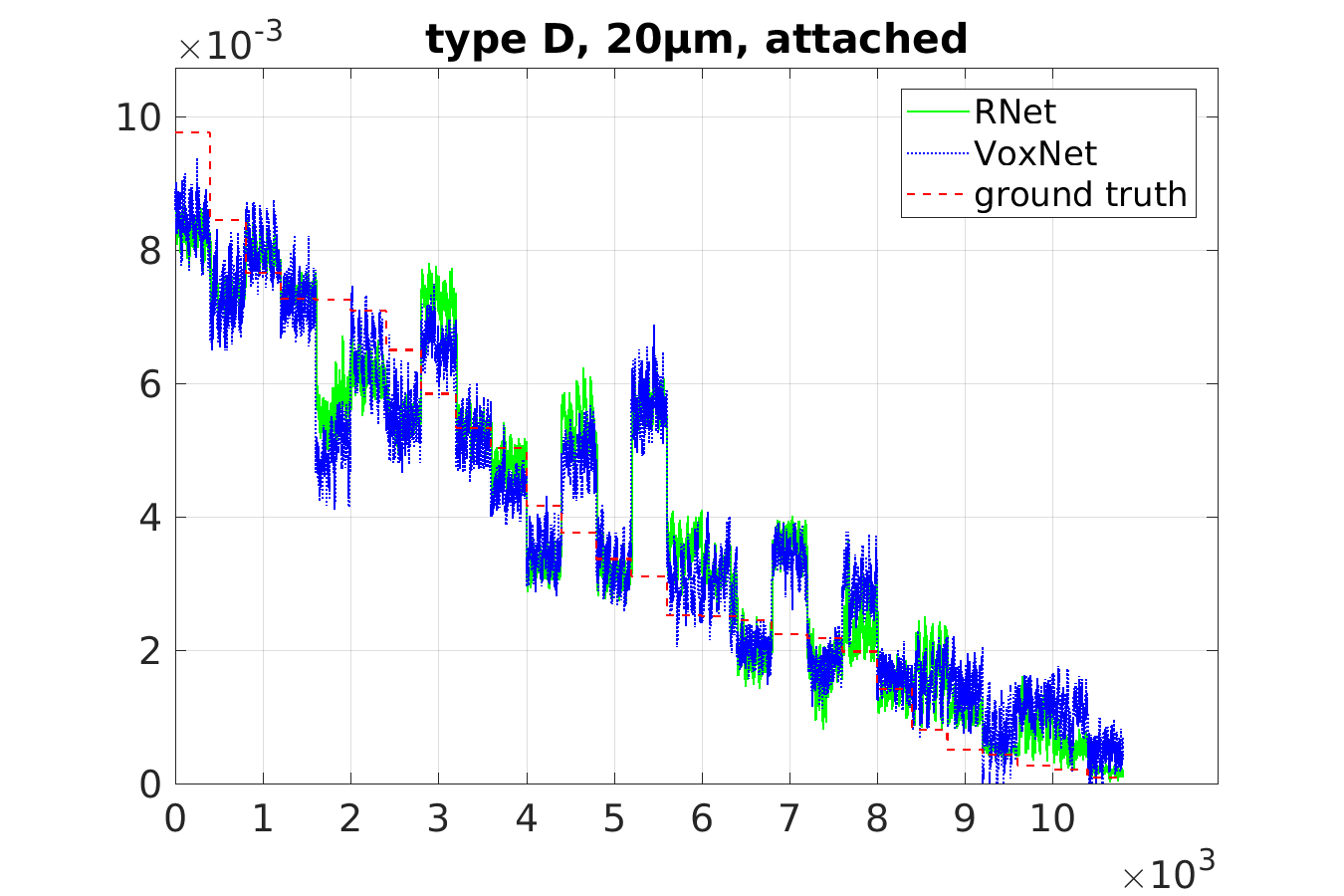}}
\hfil
\subfloat{\includegraphics[width=0.2\linewidth]{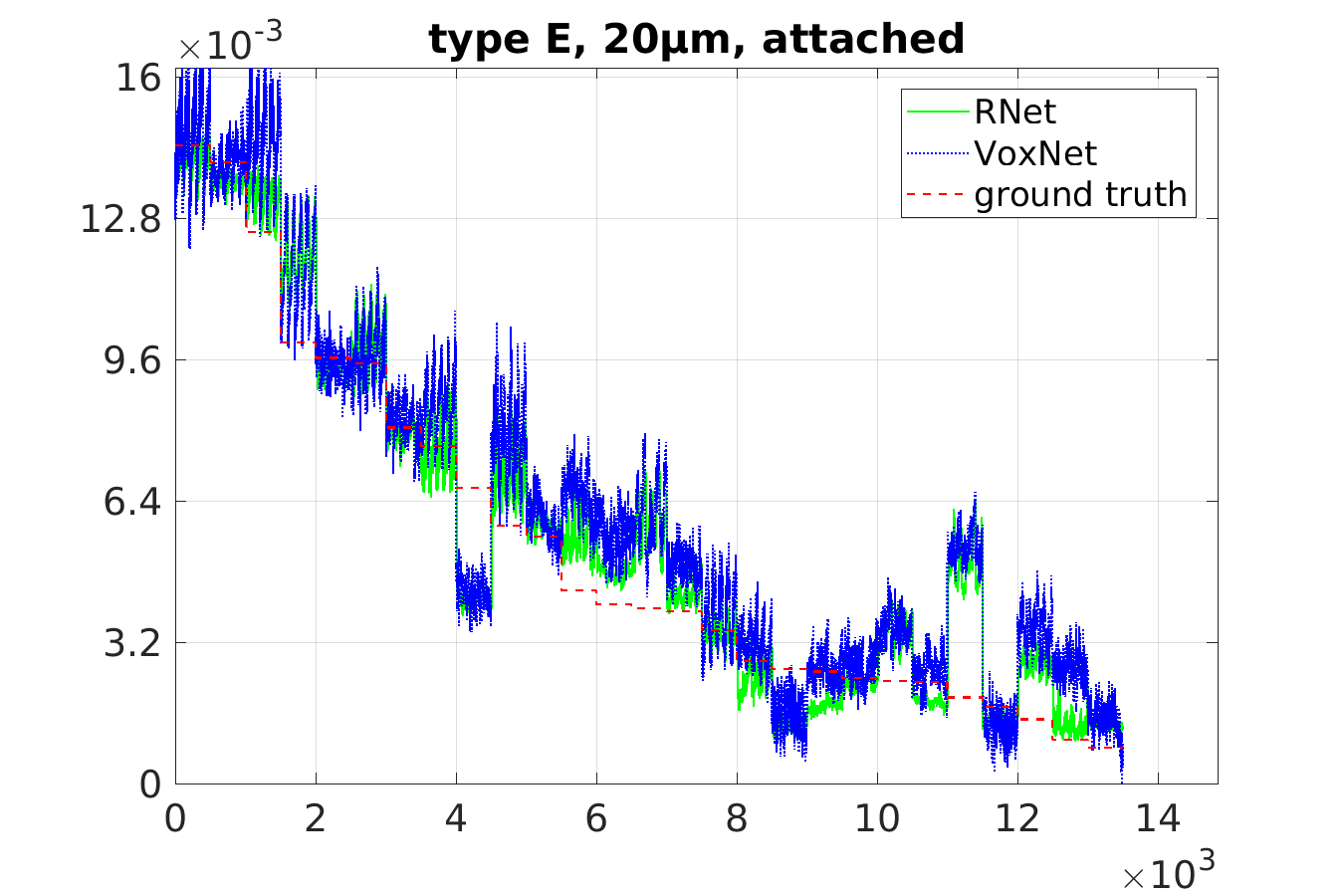}}
\hfil
\subfloat{\includegraphics[width=0.2\linewidth]{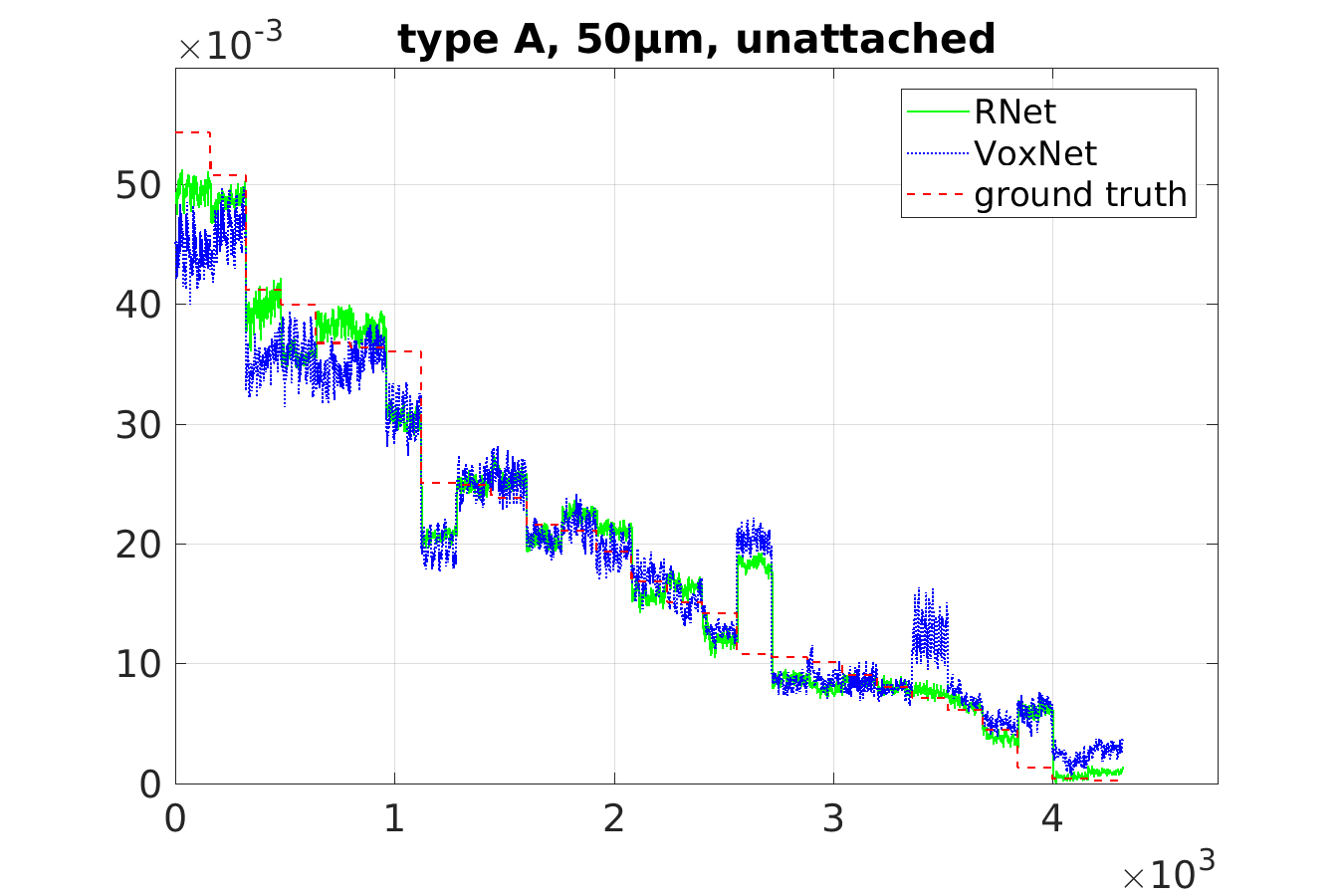}}
\hfil
\subfloat{\includegraphics[width=0.2\linewidth]{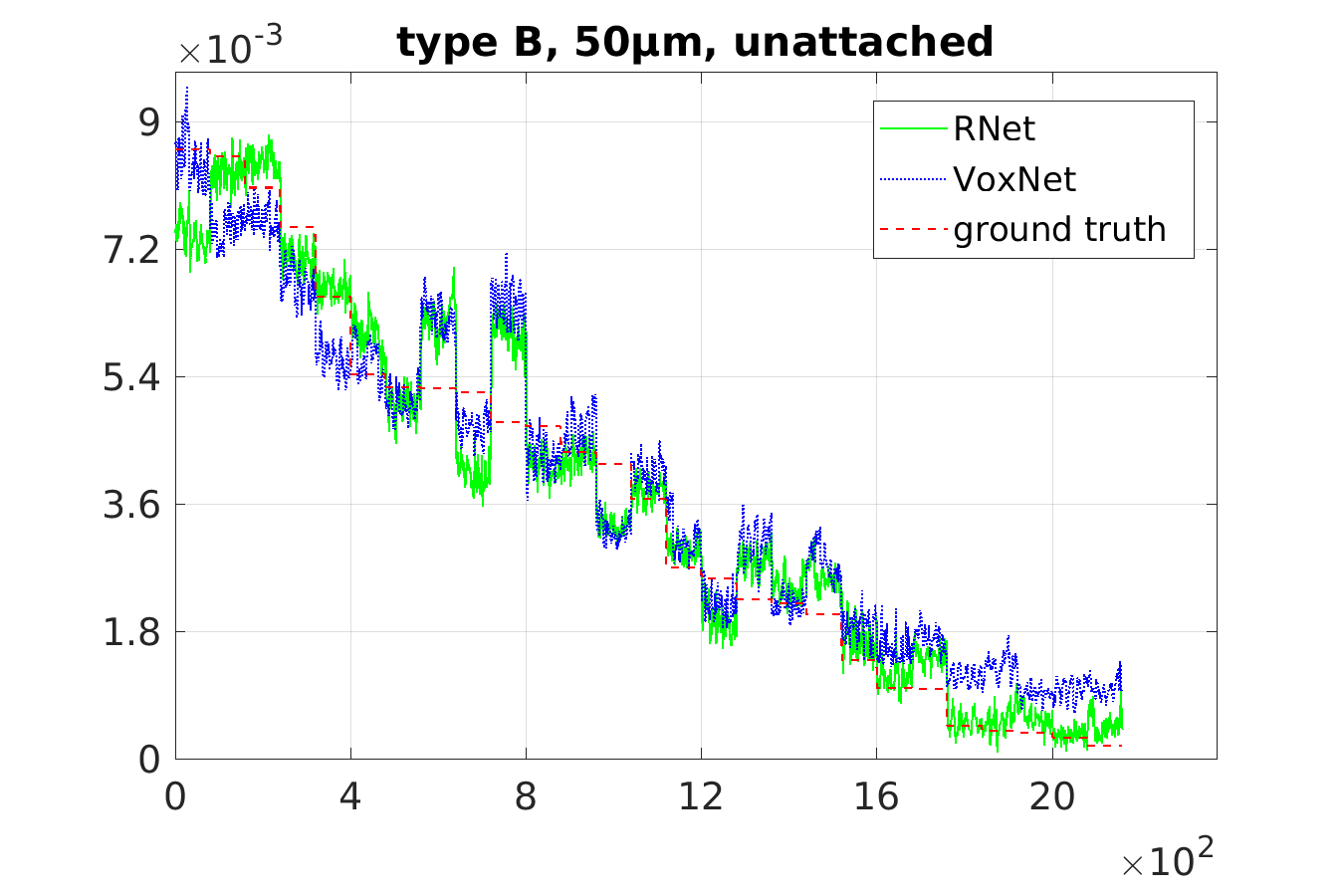}}
\hfil
\subfloat{\includegraphics[width=0.2\linewidth]{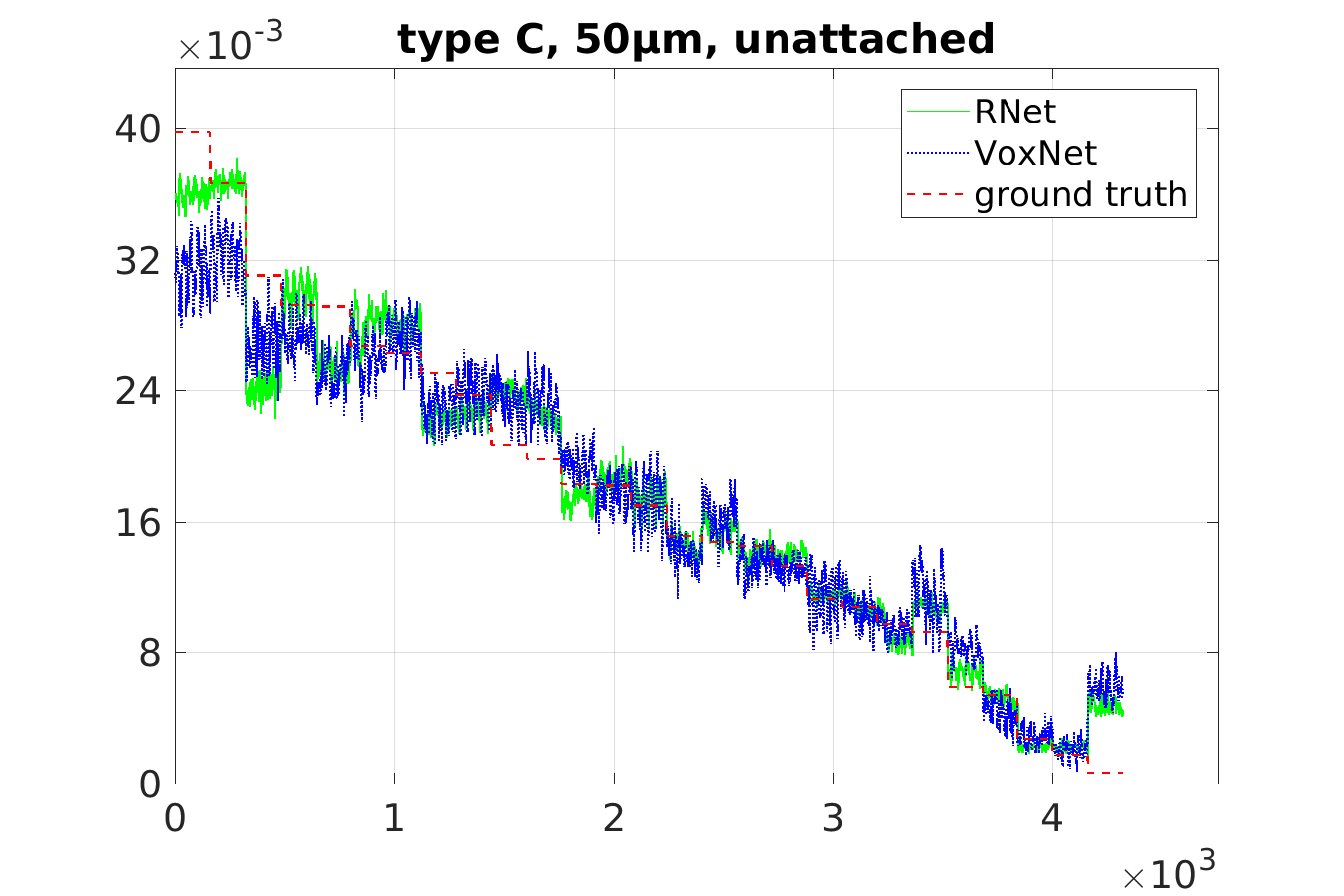}}
\hfil
\subfloat{\includegraphics[width=0.2\linewidth]{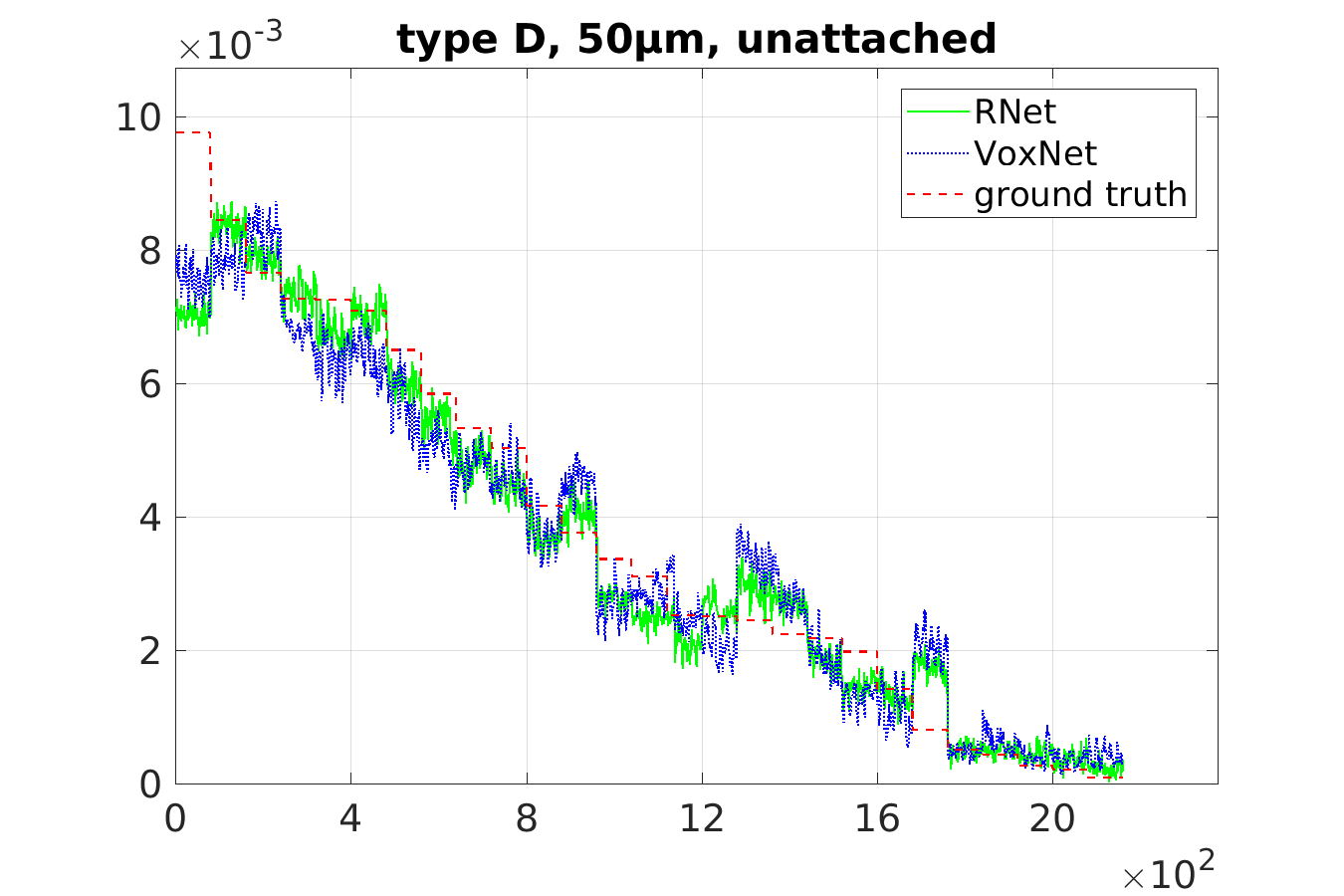}}
\hfil
\subfloat{\includegraphics[width=0.2\linewidth]{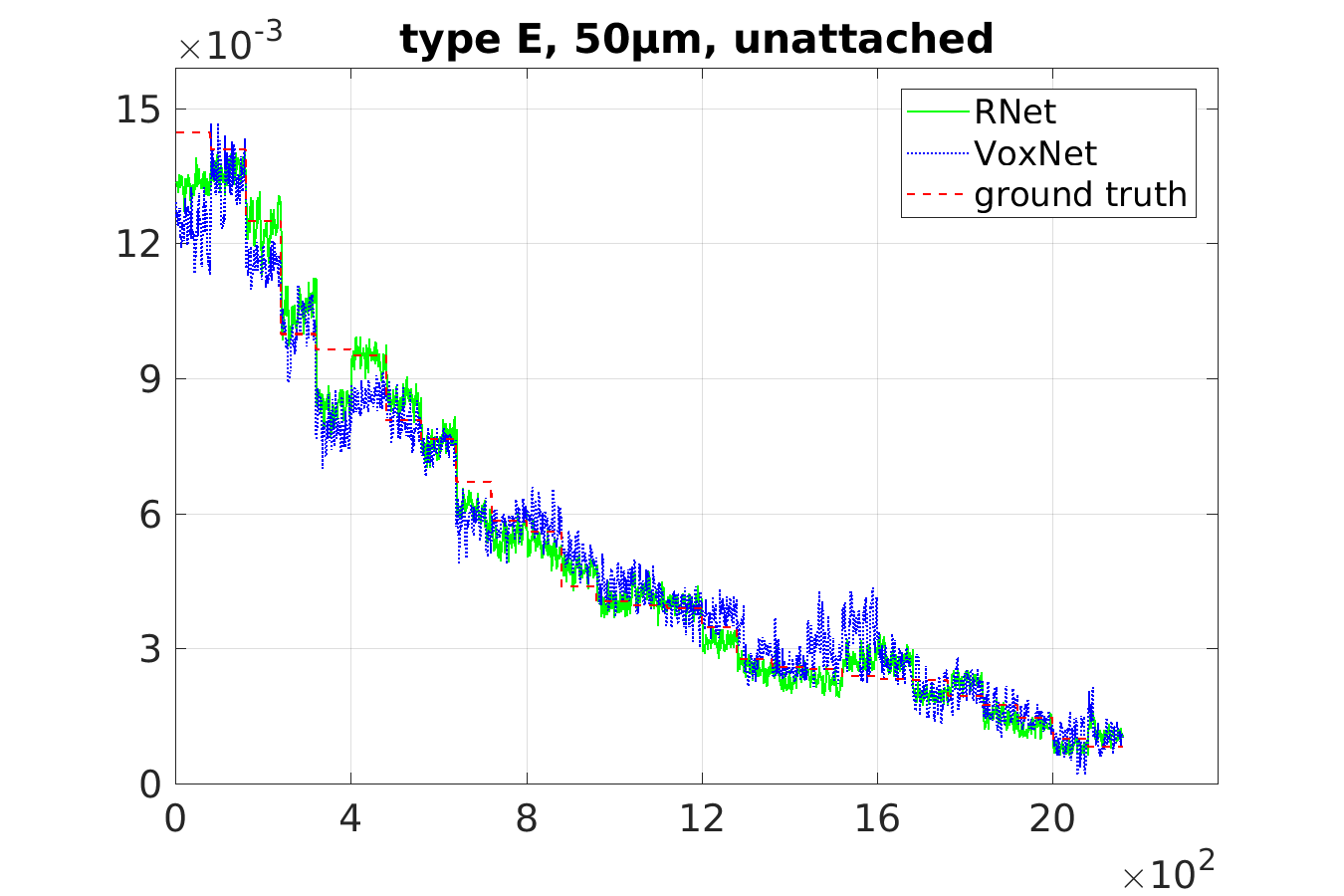}}
\hfil
\subfloat{\includegraphics[width=0.2\linewidth]{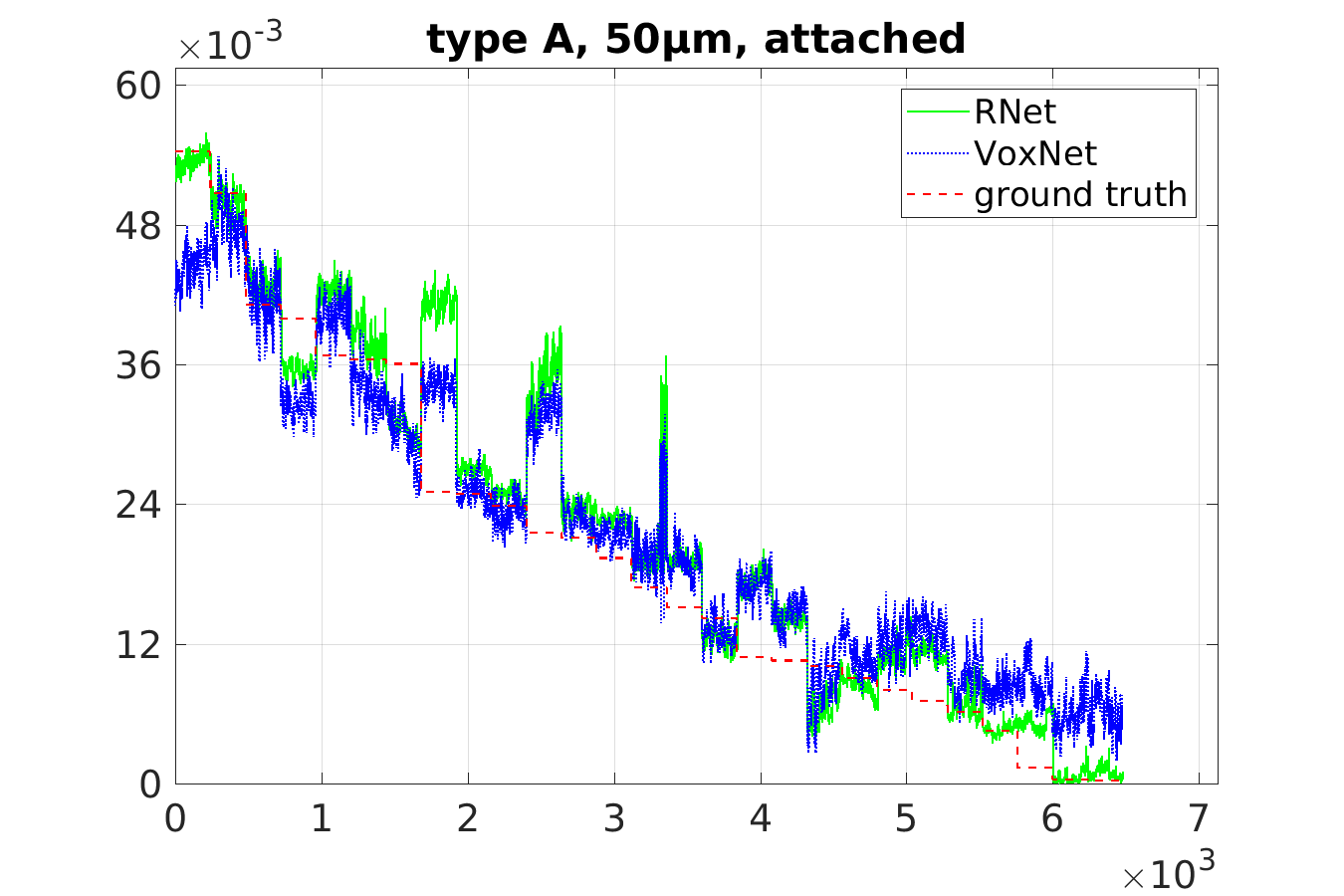}}
\hfil
\subfloat{\includegraphics[width=0.2\linewidth]{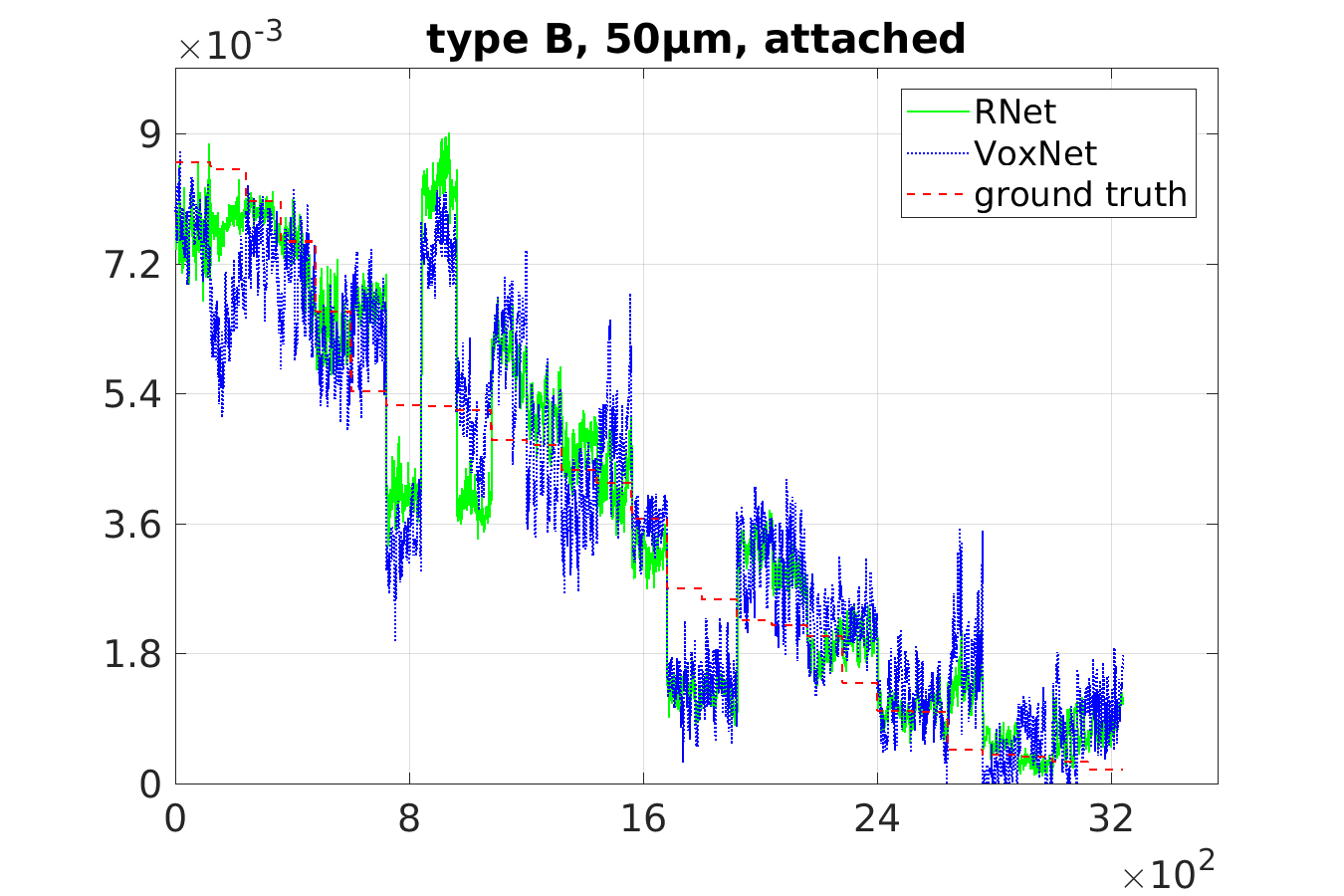}}
\hfil
\subfloat{\includegraphics[width=0.2\linewidth]{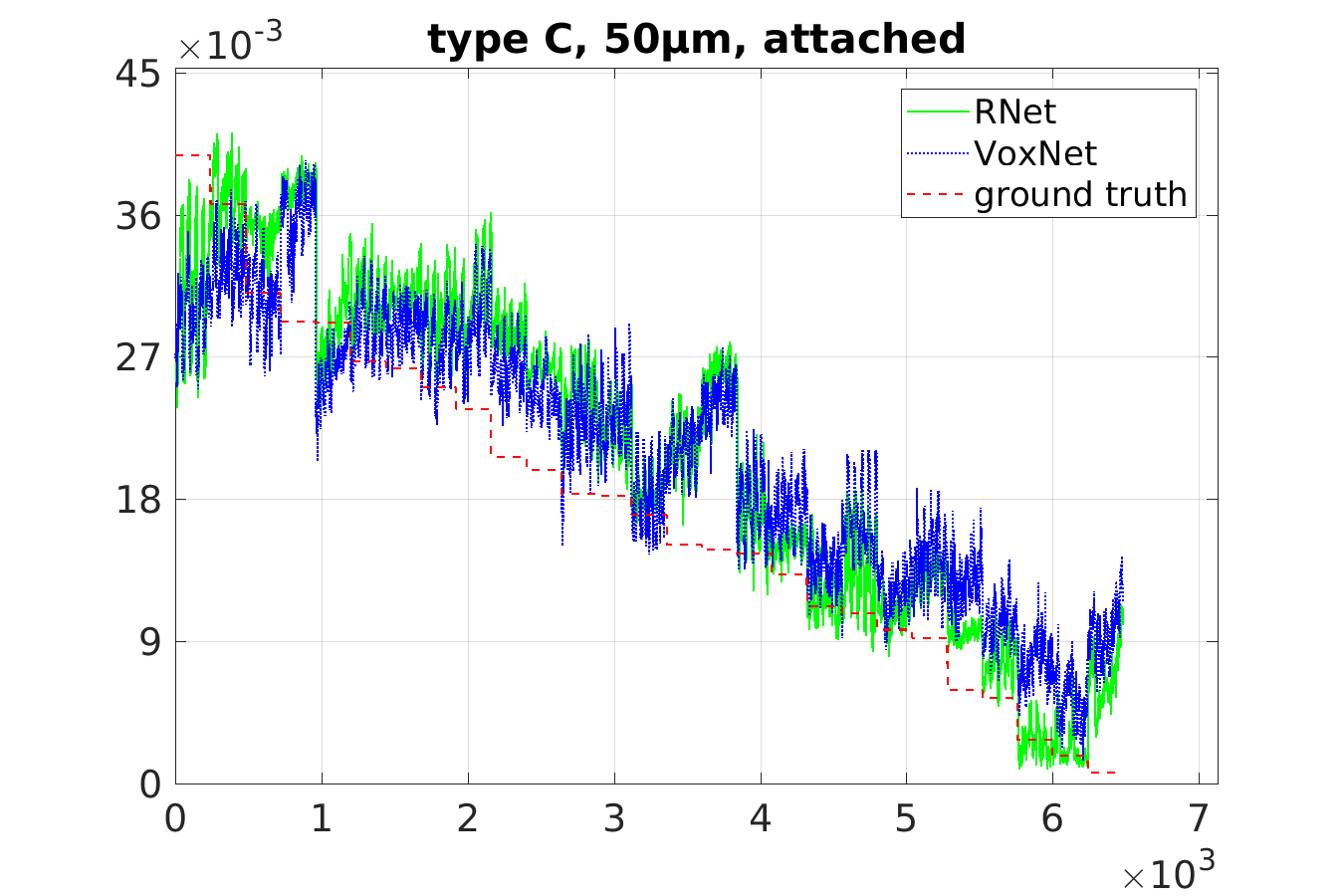}}
\hfil
\subfloat{\includegraphics[width=0.2\linewidth]{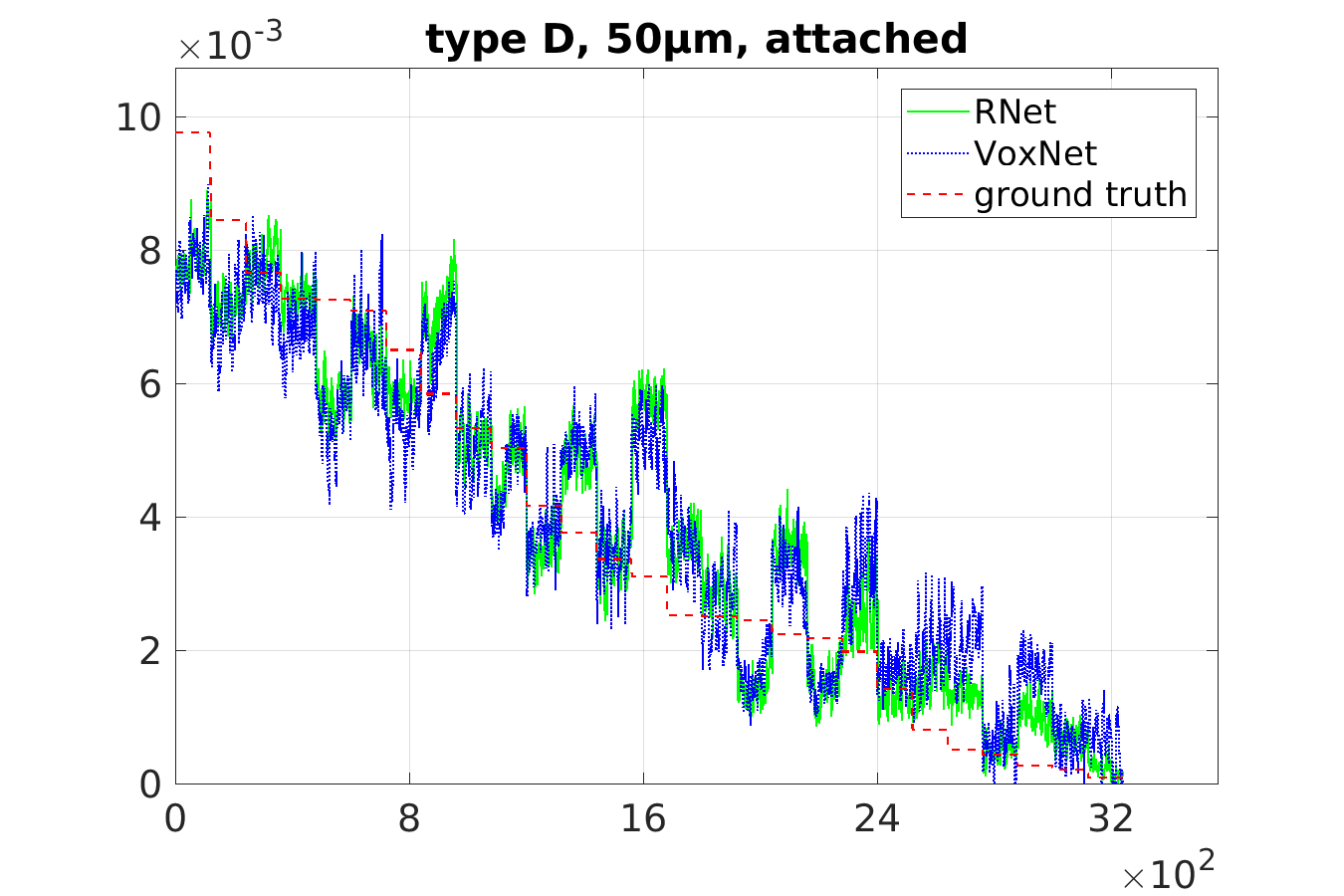}}
\hfil
\subfloat{\includegraphics[width=0.2\linewidth]{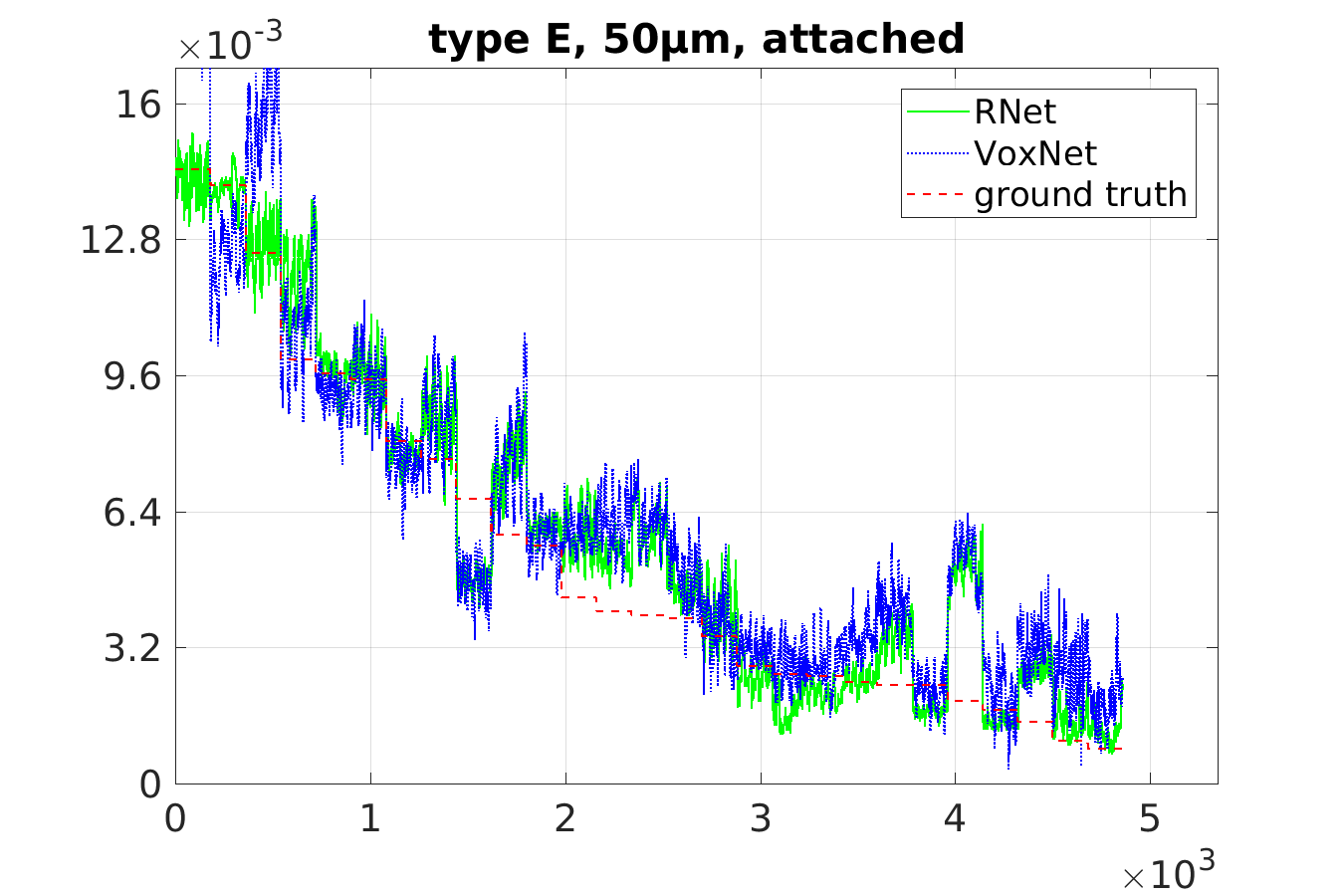}}
\caption{Comparison of 3DCNN predictions and ground truth. The vertical axis corresponds to the glue volume and the horizontal axis to the index of the test sample. For each model testing samples are sorted according to the ground truth glue volume, and results for sampling steps of $20{\mu}m$ and $50{\mu}m$ are reported. Even in the more challenging cases where dies have been attached, the predictions of the 3DCNN models (in green) follow the ground truth curves (in red).}
\label{fig:predresults}
\end{figure*}

\begin{table}
\small
\centering
\begin{tabular}{|c|c||c|c|c|c|c|}
\hline
\multicolumn{2}{|c|}{\textit{Type}} & \textit{A} & \textit{B} & \textit{C} & \textit{D} & \textit{E}  \\
\hline
\hline
\multicolumn{7}{|c|}{\textbf{20 micrometer sampling step}} \\
\hline 
\multirow{2}{*}{\textit{no IC}} & RNet & \textit{8.21} & \textit{0.45} & \textit{5.37} & \textit{0.39} & \textit{0.21} \\
\cline{2-7}
 & VoxNet & 29.55 & 0.67 & 9.89 & 0.71 & 0.38 \\
\hline
\multirow{2}{*}{\textit{IC}} & RNet & \textit{21.54} & 1.25 & \textit{25.19} & \textit{1.02} & \textit{1.37} \\
\cline{2-7}
 & VoxNet & 28.46 & \textit{1.07} & 25.28 & 1.04 & 2.26  \\
\hline
\multicolumn{7}{|c|}{\textbf{50 micrometer sampling step}} \\
\hline
\multirow{2}{*}{\textit{no IC}} & RNet & \textit{7.99} & \textit{0.35} & \textit{5.29} & \textit{0.47} & \textit{0.25} \\
\cline{2-7}
 & VoxNet & 16.05 & 0.54 & 8.71 & 0.58 & 0.60 \\
\hline
\multirow{2}{*}{\textit{IC}} & RNet & \textit{26.89} & \textit{1.00} & \textit{28.40} & \textit{1.00} & \textit{1.55} \\
\cline{2-7}
 & VoxNet & 28.12 & 1.21 & 30.72 & 1.29 & 3.07 \\
\hline
\end{tabular}
\caption{Validation MSE in $e{-06}$ $(mm^3)^2$ for RNet and VoxNet.}
\label{tab::res}
\end{table}

\begin{table}
\small
\centering
\begin{tabular}{|c|c||c|c|c|c|c|}
\hline
\multicolumn{2}{|c|}{\textit{Type}} & \textit{A} & \textit{B} & \textit{C} & \textit{D} & \textit{E}  \\
\hline
\hline
\multicolumn{7}{|c|}{\textbf{20 micrometer sampling step}} \\
\hline 
\textit{no IC} & PointNet & 2.343 & 33.968 & 0.618 & 1.623 & 0.800 \\
\hline
\textit{IC} & PointNet & 39.57 & 0.265 & 22.025 & 0.094 & 1.840 \\
\hline
\end{tabular}
\caption{Validation MSE in $e{-03}$ $(mm^3)^2$  for PointNet.}
\label{tab::res-pnet}
\end{table}

Although the volume estimation task that we focus on can have a wider industrial applicability, to the best of our knowledge it has not been studied before. Thus there are no obvious alternative deep architectures that we can compare with. To this end, we have adjusted two well known methods, namely VoxNet \cite{matu0} and PointNet \cite{pointnet}, to our volume estimation task. For VoxNet we have replaced its final classification layer with a fully connected layer in order to produce a single volume estimation whereas for PointNet we followed the same procedure for its classification sub-network. Also since PointNet takes as input $1024$ 3D points we have uniformly subsampled each of our $20{\mu}m$ 3D scans to sample this number of points.

The MSE over the different validation sets is reported in Tables \ref{tab::res} and \ref{tab::res-pnet}. Notice that the results for PointNet (Table \ref{tab::res-pnet})  are much worse compared to VoxNet and RNet (Table \ref{tab::res}). Although PointNet has shown excellent results for 3D object classification, it performs poorly in the current task and its architecture needs to be further altered. Probably in its current form it cannot encode accurately the subtle differences between different quantities of glue from direct point cloud input. RNet outperforms VoxNet as well in almost all experiments.

\begin{table}
\small
\centering
\begin{tabular}{|c|c|c|c|}
\hline
\textit{Scanning Step} & \textit{Scanning Time} & \textit{Prediction Time} & \textit{Total Time}  \\
\hline
\hline 
\textit{$20{\mu}m$} & 2620 & 470 & \textbf{3090} \\
\hline
\textit{$50{\mu}m$} & 1181 & 470 & \textbf{1651} \\
\hline
\end{tabular}
\caption{PCB Total Scanning \& Inspection Time in $secs$.}
\label{tab::perform-time}
\end{table}

As is anticipated the error is significantly larger after dies have been attached since glue mass can not be directly scanned and inevitably there are discrepancies in the ground truth values as explained in Section \ref{sec::dataset}. Nonetheless, test error is not substantially affected by having a larger sampling step, showcasing the ability of RNet and to a lesser degree VoxNet to infer the volume of glue even from a sparser point cloud. Having a larger sampling step is important for accelerating the scanning process and not delaying the PCB manufacturing process. The required scanning and prediction time for a single PCB with a step of $20{\mu}m$ or $50{\mu}m$ is shown in Table \ref{tab::perform-time}.  During prediction point clouds are examined one by one and not in a batch allowing an earlier notification in case of a defect. In both cases the total time is below the critical time threshold of one hour after which glue viscosity drops and any repair activity becomes much more difficult.

While Tables \ref{tab::res}, \ref{tab::res-pnet} and \ref{tab::perform-time} are useful for comparing different experimental setups, from the end-user's point of view it is more important to examine the ability of the trained models to discern the fluctuations of glue quantity. In this regard, Figure \ref{fig:predresults} shows the performance of VoxNet and RNet for each test set in comparison with the ground truth. The vertical axis refers to the volume of glue, while the lateral one to the index of the test set sample. On each plot the red curve shows the ground truth glue volume while the green and blue curves the predictions of RNet and VoxNet respectively. We have excluded PointNet from this comparison since its high MSE would hinder visualization. Each test set has been sorted according to the annotated quantity of glue in the respective samples, thus the red curves are monotonically decreasing. The diagrams of Figure \ref{fig:predresults} verify Table \ref{tab::res} since there is no substantial difference when increasing the scanning step while accuracy drops after die attachment. Also RNet obviously follows more tightly ground truth in several instances (e.g. type E unattached) whereas in other cases is covered by VoxNet due to VoxNet's wider fluctuations. However in general, the trained models can indeed capture the different levels of glue since their predictions follow the ground truth curves. Of course when dies are attached there are stronger deviations from the ground truth, especially for type B and D glue depositions where the scanned area is smaller and deposited quantity of glue is less. Nonetheless the general downward trend is efficiently captured in all cases. Even after die attachment RNet can infer glue volume from the exceeding glue around the die and from the elevation of the die above the LCP substrate due to the intermediate glue layer, therefore providing meaningful predictions for fault diagnosis.

\subsection{Fault diagnosis evaluation}
\label{sec::exclass}

\begin{figure}
\centering
\subfloat[]{\includegraphics[width=0.95\linewidth]{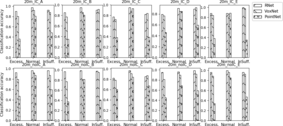}\label{fig:classresultsA}}
\hfil
\subfloat[]{\includegraphics[width=0.95\linewidth]{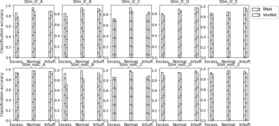}\label{fig:classresultsB}}
\caption{Classification results for RNet, VoxNet and PointNet. RNet outperforms the other architectures in almost all cases.}
\label{fig:classresults}
\end{figure}

Volume estimations can be used for automated fault diagnosis based on predefined threshold of normal quantities. To this end, using expert knowledge we have defined such threshold and evaluated the performance of RNet, VoxNet and PointNet for all experiment scenarios. In Figures \ref{fig:classresultsA} and \ref{fig:classresultsB} we see the classification accuracy for RNet using a $20{\mu}m$ or $50{\mu}m$ scanning step. Again RNet outperforms VoxNet and PointNet with an average classification accuracy of $91.82$ \% compared to $86.42$ \% and $58.33$ \% respectively.

\section{Conclusions}
\label{sec:conclusions}

In this paper a fault diagnosis system is proposed for inspecting glue dispensation and die attachment on PCBs. To realize this, a custom scanning module has been built using a commercially available laser scanner and two high-accuracy linear stages while a hardware triggering mechanism has been implemented in order to accelerate the scanning process. This paper also introduces RNet architecture in order to monitor a geometric quantity, namely the volume of glue from the scanned point clouds corresponding to different regions of interest on a PCB. To train and deploy the 3DCNN data parameterization and augmentation steps have been introduced.

The efficiency of the system is demonstrated through several experiments on different types of glue regions. A main finding of our work is that glue volume can be estimated even after dies have been attached and most of the glue is occluded. Although the error increases in this case, the predictions of RNet are still useful for quality control. This enables the automatic inspection at a later stage, after dies are attached thus postponing a time consuming scanning process that might affect the viscosity of glue especially in even more complex PCBs. Another interesting outcome of our research is that prediction accuracy is not significantly affected by a larger scanning step as similar performance is noticed for sparser point clouds at a significantly reduced scanning time.

As a next step it would be interesting to deploy and evaluate the proposed framework in other industrial use cases where a geometric parameter needs to be monitored and further research challenges would arise that have not been addressed in the current work. Such challenges could be related to the misalignment of the sensor and specimen that could result in partial scans and the detection of multi-scale defects that would require adjustments in both data parameterization and network architecture.

%\section*{Acknowledgment}
%
%This work has been partially supported by the European Commission through project Z-Fact0r funded by the European Union H2020 programme under Grant Agreement no. 723906. The opinions expressed in this paper are those of the authors and do not necessarily reflect the views of the European Commission.

% References
%
\bibliographystyle{Bibliography/IEEEtranTIE}
\bibliography{Bibliography/TIE2019}\ %IEEEabrv instead of IEEEfull
%\bibliography{Bibliography/IEEEabrv,Bibliography/TIE2019}\ %IEEEabrv instead of IEEEfull

%\vspace{-1cm}
\begin{IEEEbiography}[{\includegraphics[width=1in,height=1.25in,clip,keepaspectratio]{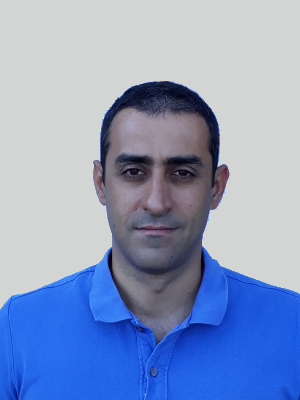}}]{Nikolaos Dimitriou}
received the diploma in electical and computer engineering from the Aristotle University of Thessaloniki (AUTh) in 2007. From the same institution he received the PhD degree in 2014. Since the July of 2014, he works a postdoctoral research associate for the Informatics and Telematics Institute   of the Centre for Research and Technology Hellas in Thessaloniki. His research focuses on the areas of machine vision and deep learning with a focus on industrial applications.
\end{IEEEbiography}

%\vspace{-2cm}
\begin{IEEEbiography}[{\includegraphics[width=1in,height=1.25in,clip,keepaspectratio]{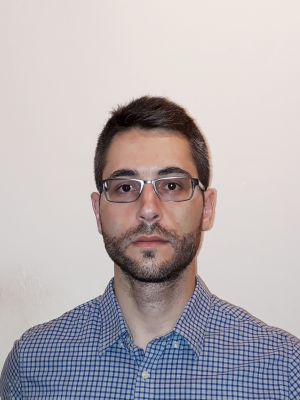}}]{Lampros Leontaris}
received the diploma in electical and computer engineering from the Aristotle University of Thessaloniki (AUTh) in 2016. He works as a research assistant at the Information Technologies Institute of the Centre for Research and Technology Hellas in Thessaloniki (CERTH/ITI) since March 2017.His research interests include computational intelligence methods, machine learning and fuzzy systems.
\end{IEEEbiography}

%\vspace{-2cm}
\begin{IEEEbiography}[{\includegraphics[width=1in,height=1.25in,clip,keepaspectratio]{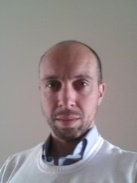}}]{Thanasis Vafeiadis}
received his diploma in mathematics in 2002, master?s degree in Statistics and Operation Research in 2004 and PhD in Departmental Investigation of Linear Trend in Time Series in 2011, from Aristotle University of Thessaloniki (AUTh). He is currently working as postdoctoral research associate in Informatics and Telematics Institute of the Centre for Research and Technology Hellas in Thessaloniki. His main research interests focus on the analysis of linear trend in time series, in machine learning, signal processing and data analytics.
\end{IEEEbiography}

%\vspace{-2cm}
\begin{IEEEbiography}[{\includegraphics[width=1in,height=1.25in,clip,keepaspectratio]{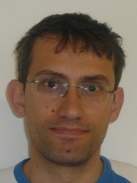}}]{Dimosthenis Ioannidis}
received his diploma in electrical and computing engineering in 2000 master?s degree in advanced communication systems and engineering in 2005 from Aristotle University of Thessaloniki (AUTh). He is working from October 2006 till now as a Research Associate in Informatics and Telematics Institute of the Centre for Research and Technology Hellas in Thessaloniki. His main research interests are in the areas of biometrics (gait and activity-related recognition), 3-D data processing, and web semantics.
\end{IEEEbiography}

%\vspace{-2cm}
\begin{IEEEbiography}[{\includegraphics[width=1in,height=1.25in,clip,keepaspectratio]{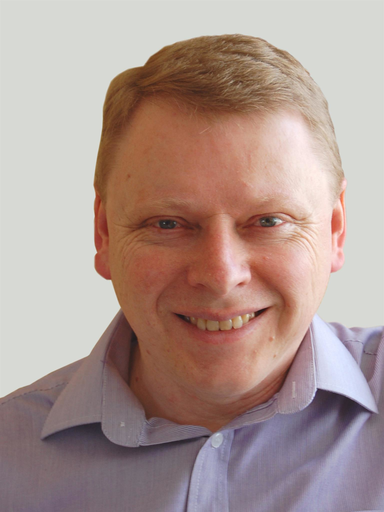}}]{Tracy Wotherspoon}
graduated from Birmingham City University in 1984 with a HND in Electrical and Electronic Engineering. Following graduation he worked in the process control industry before transferring to the design of broadcast quality sound mixers with Audio Developments (Walsall) and latterly Coomber Electronic Equipment (Worcester). Following Microsemi's, then Zarlink's, move into medical electronic assembly he was awarded an MSc in Clinical Engineering from Cardiff University specialising in medical RF. He has patents in the field of energy harvesting, as well as several publications and book chapters in medical implant communications and harsh environment electronics.
\end{IEEEbiography}

%\vspace{-2cm}
\begin{IEEEbiography}[{\includegraphics[width=1in,height=1.25in,clip,keepaspectratio]{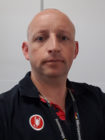}}]
{Gregory Tinker}
graduated with a HNC in Electronics and Communications from Plymouth University in 1993. Following graduation he worked at Meggitt electronics (Holsworthy) as a production engineer until 1998 and subsequently moved to Mitel (south Wales) as an equipment engineer. The company has changed name several times since and so have the roles he has fulfilled in 20 years of service including SMT process engineer, equipment and facilities senior engineer and currently holds the position of senior process engineer.
\end{IEEEbiography}

%\vspace{-2cm}
\begin{IEEEbiography}[{\includegraphics[width=1in,height=1.25in,clip,keepaspectratio]{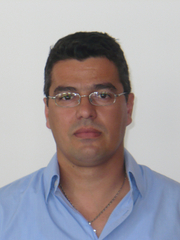}}]{Dimitrios Tzovaras}
(SM13) received the Diploma and the Ph.D. degree in electrical and computer engineering from Aristotle University of Thessaloniki, Thessaloniki, Greece, in 1992 and 1997, respectively. He is the Director at the Information Technologies Institute of the Centre for Research and Technology Hellas. His main research interests include visual analytics, 3D object recognition, search and retrieval, behavioral biometrics, assistive technologies, information and knowledge management, computer graphics, and virtual reality.
\end{IEEEbiography}

\end{document}